%% file: main.tex
\documentclass[10pt,emptycopyrightspace]{ewsn-proc}

\usepackage{graphicx}
\usepackage{balance}
\usepackage{comment}

\usepackage{amsmath}
\usepackage{amssymb}
\usepackage{xcolor}
\usepackage{xspace}
\usepackage{caption}
\usepackage{subcaption}
\usepackage{multirow}
\usepackage{makecell}

\usepackage{tabularray} % for table1
\usepackage{threeparttable} % footnotes for table
\usepackage{tablefootnote}  % footnotes for table

\usepackage{url} % for youtube link
\usepackage[colorlinks, linkcolor=black, anchorcolor=red, citecolor=black]{hyperref}
\newcommand{\sysname}{\texttt{UTM}\xspace}
\usepackage[misc,geometry]{ifsym}
\author{Kaiwen Cai$^{1*}$, Qiyue Xia$^{2*}$, Peize Li$^2$, John Stankovic$^3$, Chris Xiaoxuan Lu$^2$\Letter
\\\affaddr{$^*$Co-primary authors, \Letter\ Corresponding author}
\\\affaddr{$^1$University of Liverpool, $^2$University of Edinburgh, $^3$University of Virginia}
\\\email{$^1$k.cai@liverpool.ac.uk, $^2$xqy170605@gmail.com, \{peize.li, xiaoxuan.lu\}@ed.ac.uk, $^3$stankovic@cs.virginia.edu}
}

\title{Robust Human Detection under Visual Degradation \\ via Thermal and mmWave Radar Fusion}

\begin{document}

\maketitle

\input{sections/0_abstract}

\input{sections/1_introduction.tex}
\input{sections/2_prelimineries.tex}
\input{sections/4_methods.tex}
\input{sections/5_experiments.tex}
\input{sections/6_related_work.tex}
\input{sections/7_conclusion.tex}

%
% NOTE
%
% The following commands are all you need in the
\balance
\bibliographystyle{abbrv}
\bibliography{sigproc}  % sigproc.bib is the name of the Bibliography in this case
\end{document}

%% file: sections/0_abstract.tex
\begin{abstract}
    The majority of human detection methods rely on the sensor using visible lights (e.g., RGB cameras) but such sensors are limited in scenarios with degraded vision conditions. In this paper, we present a multimodal human detection system that combines portable thermal cameras and single-chip mmWave radars. To mitigate the noisy detection features caused by the low contrast of thermal cameras and the multi-path noise of radar point clouds, we propose a Bayesian feature extractor and a novel uncertainty-guided fusion method that surpasses a variety of competing methods, either single-modal or multi-modal. We evaluate the proposed method on real-world data collection and demonstrate that our approach outperforms the state-of-the-art methods by a large margin.
\end{abstract}

%% file: sections/1_introduction.tex
\begin{figure}[!t]
    \centering
    \begin{subfigure}[t]{0.9\linewidth}
        \centering
        \includegraphics[width=\linewidth]{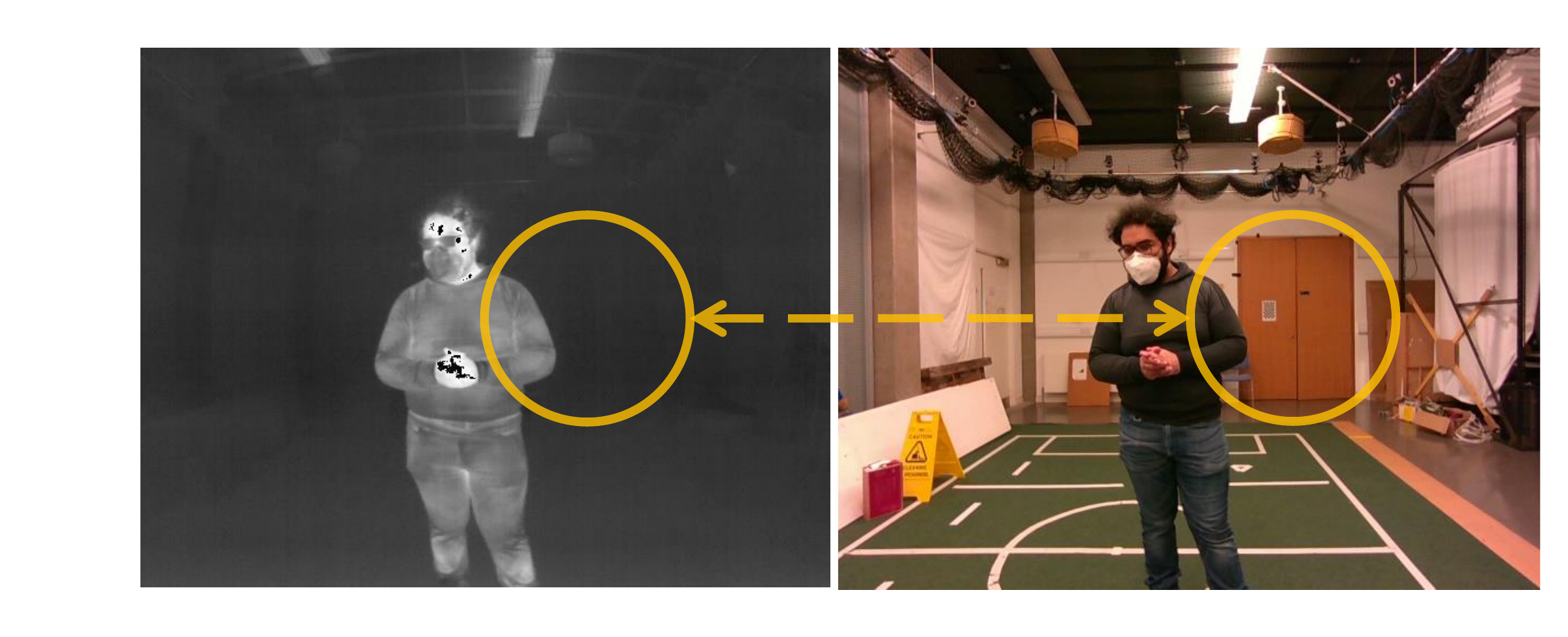}
        \caption{Thermal image (left) provides much fewer details than RGB image (right). }
        \label{fig:thermal_detail}
    \end{subfigure}
    \begin{subfigure}[t]{0.9\linewidth}
        \centering
        \includegraphics[width=\linewidth]{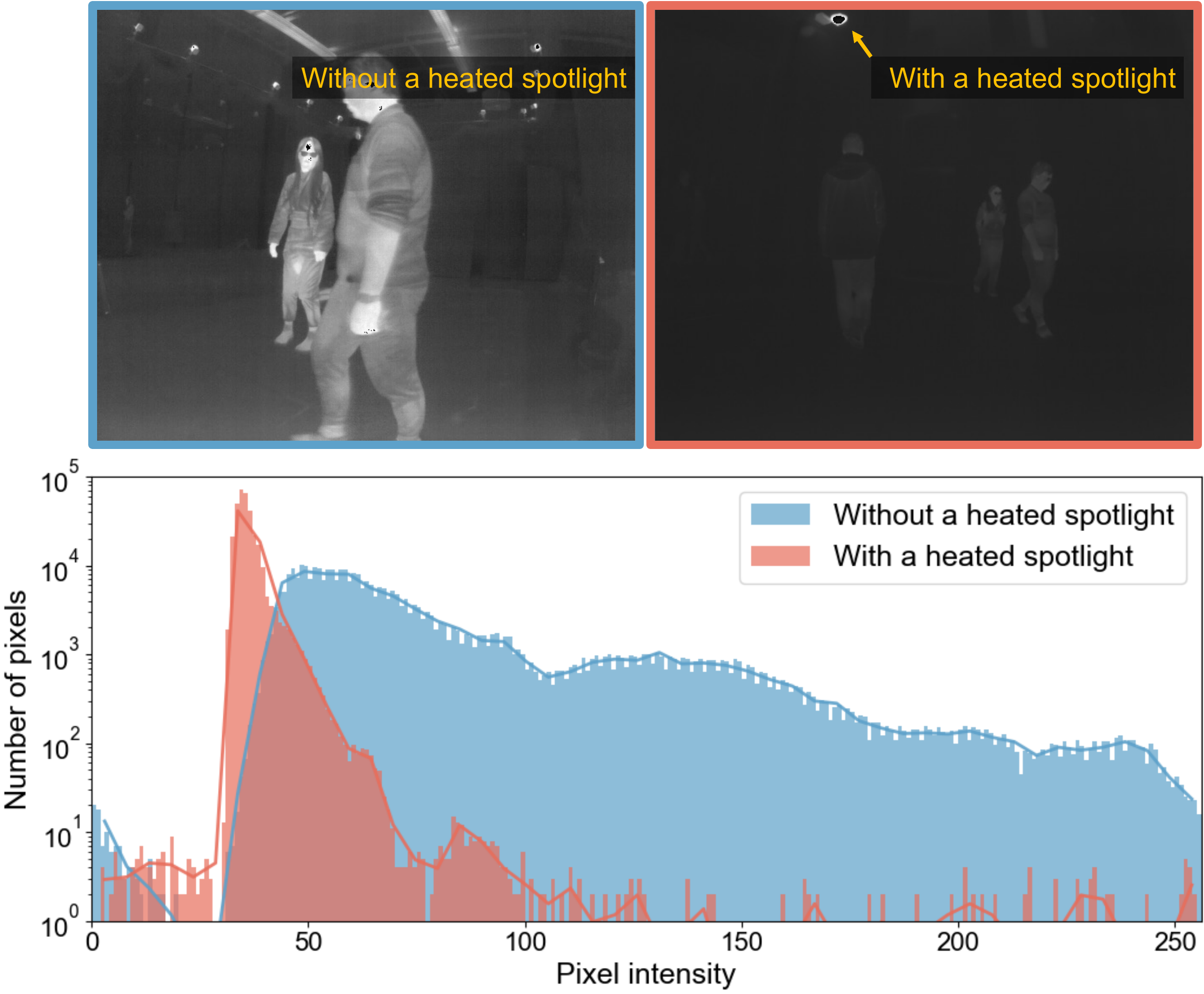}
        \caption{A heated spotlight in the background decreases the contrast of the thermal image.}
        \label{fig:thermal_contrast}
    \end{subfigure}    
    \caption{Limitation of thermal cameras.}
    \label{fig:thermal_imaging}
    \vspace{-0.5cm}
\end{figure}

\section{Introduction}
% Importance of human detection

Human detection is the task of locating all instances of human beings present in sensor observations. An accurate human detection module is widely regarded as an essential component in human-centred computing and cyber-physical systems. More often than not, detecting human subjects is the precursor step to enable 
subsequent context inference, such as pose estimation, occupancy monitoring, human tracking and crowd activity recognition etc. 

\begin{figure}[h]
    \centering
    \includegraphics[width=\linewidth]{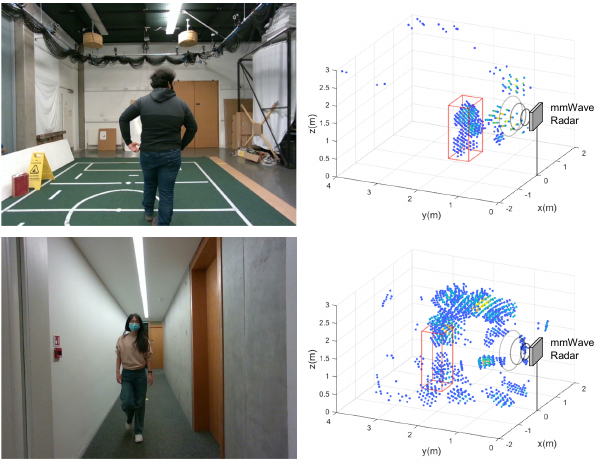}
    \caption{Top row: mmWave point cloud in an open space where the multi-path effect is negligible. Bottom row: mmWave point cloud in a narrow corridor where the multi-path effect is strong. Comparing the two, we can clearly see that the multi-path effect can significantly impact the point cloud quality and human subjects (highlighted with the red boxes) are difficult to be differentiated from the background points under such effects. }
    \label{fig:mmwave_multipath}
\end{figure}

% SOTA methods for the target task and the concept gap of prior arts
Owing to their low cost and ubiquity, RGB cameras have been widely used as the \emph{de-facto} solution to human detection and achieved great performance by leveraging recent advances in deep neural networks  and computer vision. 
Unfortunately, as RGB cameras are a type of sensor operating with visible lights, the performance of RGB-camera-based methods is very susceptible to variable illumination (e.g., sun glare, dimness or darkness) and even fails under severe visual degradation, e.g., the smoke-filled fire areas and dust-filled construction sites. To address the intrinsic limitation of RGB cameras and visible light, a few works explore unconventional sensor modalities operating with invisible electromagnetic waves for human/object detection tasks. Typical examples include thermal cameras \cite{wagner2016multispectral,guo2019domain} and millimeterwave (mmWave) radars \cite{cui2022integrated,lee2020deep} - two emerging low-cost sensors that draw increasing attention from both academia and industry. As these sensors operate with electromagnetic waves of much larger wavelengths than visible light, they are fundamentally robust to variable illumination and airborne particles. However, these non-traditional sensors also suffer from their own challenges when it comes to human detection. Specifically, one well-known limitation of thermal cameras is their lack of distinctive image features or context when the temperature field is flat (see Figure~\ref{fig:thermal_detail}). Additionally, as shown in Figure~\ref{fig:thermal_contrast}, thermal images tend to have low contrast when the temperature of objects varies significantly. These two challenges jointly impose non-trivial detection challenges on thermal images even using cutting-edge computer vision algorithms. On the other hand, mmWave radars are known for their poor spatial resolution and high noise floor due to the multi-path effect, rendering themselves an unreliable sensor modality to detect humans in some indoor environments. For instance, Figure~\ref{fig:mmwave_multipath} illustrates the impact of multi-path on the mmWave radar sensing in a narrow corridor. The multi-path effect results in a cluttered point cloud with the human subjects being smeared into the background. For these reasons, the detection accuracy using thermal cameras or mmWave radars alone is far from being comparable to the one using RGB cameras in practice.  

% Our proposed fusion approach and technical challenges to solve: i. adaptive strategy? ii. resource limitations? 
Towards a more robust human detection system under various illumination and visually degraded conditions, it is intuitive to design a multimodal fusion approach that combines the strengths of thermal cameras and mmWave radars and in the meantime, compensates for their weaknesses. While such a fusion concept is straightforward, transforming it into a useful system requires addressing multiple technical challenges. First, prior arts in this vein e.g., \cite{shuai2021MilliEye} \cite{lu2020fusion} mostly adopt a late fusion strategy that individually optimises the feature extractors of each sensor modality. Such a separate feature extractor design under-exploits the cross-modal complementariness during the feature learning phase and incurs sub-optimal human detection results. Moreover, the predominant fusion operation in DNNs, e.g., the self- or cross-attention mechanisms \cite{chang2020spatial,liu2021wavoice, lu2020milliego}, generates the mask from the immediate features to weaken unimportant or noisy features. However, in our settings where the input data is sparse and noisy, the learnt attention masks could be inevitably misled by data uncertainty.
This challenge is particularly prominent in our context, as thermal or mmWave data tend to have many unimportant areas or noisy observations than RGB images.

% Method Description
In order to address these challenges, we propose \sysname 
, a novel \underline{U}ncertainty-guided \underline{T}hermal and \underline{M}mwave radar fusion framework that is able to robustly detect human subjects under visual degradation by fusing the thermal and mmWave data. \sysname follows an end-to-end optimization that leverages a Bayesian Neural Network (BNN) to extract the features from two modalities jointly and provide direct feature uncertainty to inform the mask generation. \sysname demonstrates the feasibility of thermal-mmWave fusion for human detection tasks and provides a generic framework for different host platforms (e.g., installed as a building infrastructure or embedded on headsets). 
In summary, our contributions are as follows 
\begin{enumerate}
    \item This is the first-of-its-kind work that explores the usage of portable thermal cameras and single-chip mmWave radar for robust human detection.
    \item We propose \sysname, consisting of a novel Bayesian feature extractor (BFE) and a novel uncertainty-guided fusion (UGF) method to systematically address the detection challenges of caused by thermal images and noisy radar point clouds.
    \item We evaluate our proposed \sysname on real-world data collection, and the experimental results demonstrate that our method outperforms the best of competing method by $8.4\%$ in mAP$_{50:95}$ (mean of AP$_{50}$, AP$_{65}$,.., AP$_{95}$). 
    % \chris{KW, give a number for the level of outperformance. usually using the percentage here.}
    \item The collected thermal-mmWave human detection dataset and the source code of \sysname are publicly released to the community at \url{https://github.com/ramdrop/utm}.
\end{enumerate}

%% file: sections/2_prelimineries.tex
\section{Preliminaries}

\subsection{Thermal Imaging}
\begin{figure}
    \centering
    \includegraphics[width=\linewidth]{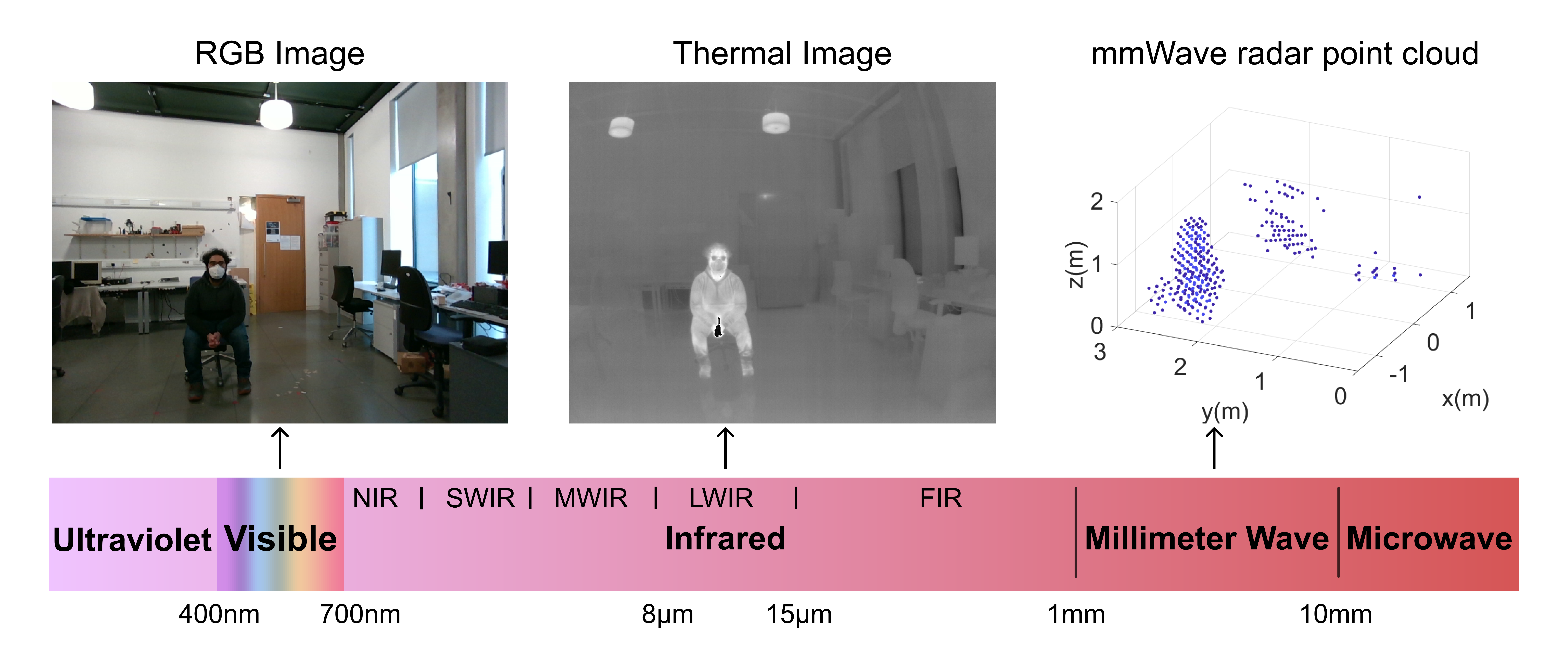}
    \caption{Electromagnetic spectrum with sub-division of infrared radiation.}
    \label{fig:spectrum}
\end{figure}

Thermal cameras \cite{narayana2020loci} capture the infrared radiation emitted by objects and convert the detected energy into pixels' values in the thermal image. The infrared radiation that can be captured lies between visible light and microwaves within the wavelength spectrum of 0.7–1,000 $\mu$m and a common sub-division scheme is illustrated in Figure~\ref{fig:spectrum}.
Among all types of thermal cameras, the long-wavelength infrared (LWIR) camera is particularly interesting and used in this work. LWIR cameras can observe the temperature field ranging from approximately 190 $\sim$ 1,000K \cite{gade2014thermal}. Importantly and similar to all other thermal cameras, LWIR cameras do not require additional sources of light or heat and can passively capture the emitted thermal energy of ambient objects. These characteristics make them a robust alternative to RGB cameras in visually degraded conditions. 

% However, compared with RGB images, the images captured by LWIR thermal cameras have less contrast and uninformative context (see  Figure~\ref{fig:thermal_detail}). Moreover, as thermal images essentially measure the differences in radiation intensity between objects and their surroundings, changes in ambient temperature can impact the quality of thermal images. This often results in thermal images lacking distinctive features as illustrated in Figure~\ref{fig:thermal_contrast}. 

% I can't see any relavance between the following paragraph and our proposed method.
% Besides the limitation brought by the working bands, thermal cameras perform Non-Uniformity Correction (NUC) which will cause periodical suspension. Most modern thermal cameras used for detection in LWIR spectral bands are based on focal plane arrays (FPA). FPAs are made up of a multitude of detector elements, where each individual detector has a different gain and offset that change with time, which will produce fixed pattern noise (FPN) in the acquired images \cite{Mudau2011Non-uniformity}. To compensate for the FPN, non-uniformity correction (NUC) must be applied, and these procedures should be performed every time the camera is powered up,  when the integration time or the lens are changed,  when the sensor or lens temperature changes, or even sometimes when the scene content changes, which will cause frequent and periodical suspension of thermal imaging. 

\subsection{Millimiter-Wave Radar}
Millimiter-Wave (mmWave) radar is a transceiver device that operates with electromagnetic waves between 30GHz-300GHz. 
mmWave radar uses a linear `chirp' or swept frequency transmission. When receiving the signal reflected by an obstacle, the radar front-end performs a dechirp operation by mixing the received signal with the transmitted signals, which produces an Intermediate Frequency (IF) signal. The distance between the object and the radar can be calculated from the IF signal.
A mmWave radar can further estimate the obstacle angle by using the different virtual antennas. Signals received at different antennas might have different phases due to the distance between the receivers. Based on the distance between the receivers and the corresponding phase differences of the received signals, the angle of arrival can be estimated \cite{peng2006angle}. While mmWave radars have been widely used in human sensing, current methods are strongly susceptible to the multi-path effect common in indoor environments. 

%% file: sections/4_methods.tex
\section{Method Design}
\begin{figure*}[t]
    \centering   \includegraphics[width=\linewidth]{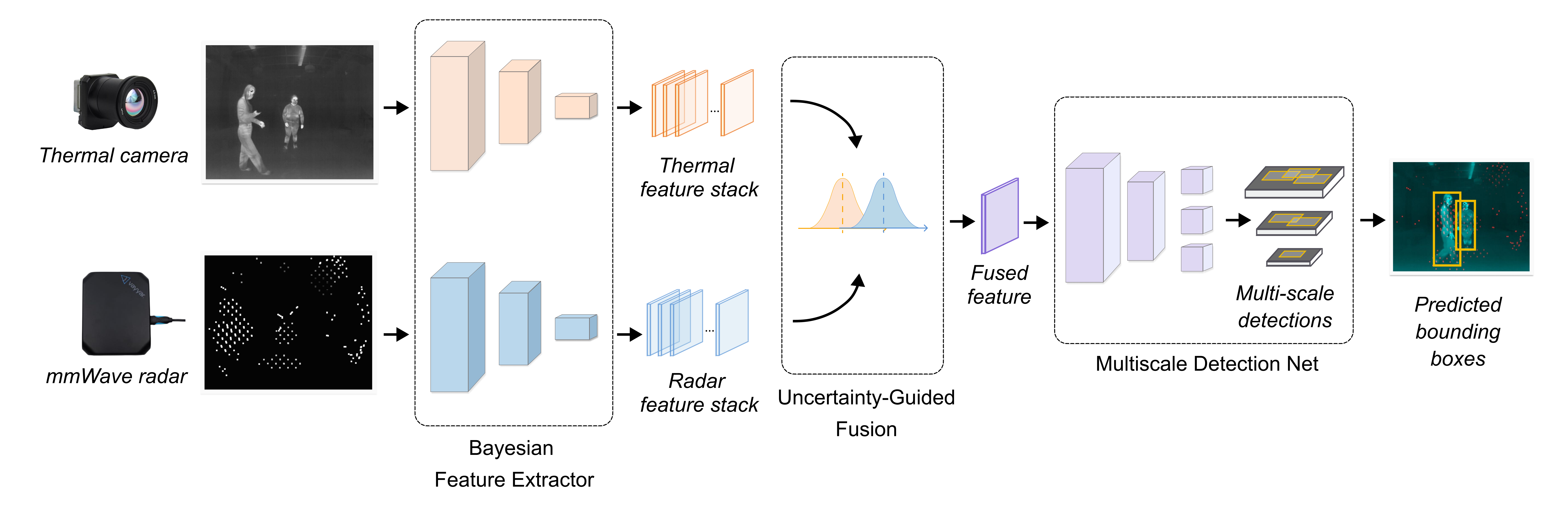}
    \caption{Overview of the proposed \sysname. }
    \label{fig_pipeline}
% \vspace{-0.5cm}    
\end{figure*}

The proposed cross-modal human detection pipeline, \sysname, is shown in Figure~\ref{fig_pipeline}. Our proposed \sysname consists of three modules: a Bayesian Feature Extractor (BFE), an Uncertainty-Guided Feature Fusion module (UGF) and a Multiscale Detection Net (MDN). Given a raw thermal image and a raw radar point cloud, we first preprocess the data to generate a radar depth image (see Sec. \ref{sec_data_preprocessing}), Next, we use the BFE module to extract features from both the thermal image and the radar depth image in parallel (see Section \ref{sec_bfe}). The features from the two modalities are then fused using the UGF module (see Section \ref{sec_ugf}). Finally, we use the MDN module to generate detection bounding boxes (see Section \ref{sec_mdn}). We detail the design of each module in subsequent sections.

% \chris{@KW, add a couple sentences to summarize the workflow of these modules. You can start with `Given a pair of radar point clouds and the thermal image, \sysname firstly preprocess them and convert ....Then, the BFE module... After that, ...'. The general principle for the first paragraph is that we need to summarize the RELATION of the following sections'}

\subsection{Data Preprocessing}
\label{sec_data_preprocessing}

Before feeding the data to the human detection network, mmWave radar and thermal camera data needs to be preprocessed. Particularly, we aim to have image-like array representation as the development of neural networks for image processing is more established and give us more design choice. Thermal images are naturally array data while mmWave radars give point clouds.
This requires us to use point cloud projection to convert the radar data into images, leading to its final representation akin to the depth images. Concretely, the radar data in each frame is a set of points, where each point is represented by a 3-$d$ vector. For clarity, we denote the point by $p:=(x,y,z)\in\mathbb{R}^3$. The mmWave radar point clouds are projected to the 2D image plane  using the extrinsic and intrinsic parameters between the thermal camera and the mmWave radar. The projection of each point follows: 
\begin{equation}
    \begin{bmatrix}
    u\\v\\1
    \end{bmatrix}
    = \frac{1}{z}KT
    \begin{bmatrix}
    x\\y\\z\\1
    \end{bmatrix}
\end{equation}
where $K$ is the $3\times3$ thermal camera's intrinsic matrix, and $T$ is the $3\times4$ extrinsic matrix between the thermal camera and the mmWave radar.  $(x,y,z)$ is the 3D location in the radar coordinate and $(u,v)$ is the projected pixel location in the 2D image plane. For each projected pixel, we use the corresponding depth value from mmWave radar point clouds to represent the pixel value on the image, which generates an image-like depth map. Meanwhile, only the projected pixels that locate within the thermal image range are reserved so that the projected depth map has the same size and field of view as the thermal image. The same size and field of view allow for more flexible fusion strategies, including the early fusion where two types of images can be directly stacked together. 
% \chris{@KW, I removed all descriptions about time sync. in the above. Please simply say that 'All sensor readings are synchronised via ROS' in the implementation section.}
% With channel concatenation, this synchronized pair of images are integrated into a 3-channel RGB-like image $(I_{thermal},I_{thermal},I_{radar\_depth\_map})$, which is a common representation for further use. 

\subsection{Bayesian Feature Extractor}
\label{sec_bfe}
% Thermal images and radar depth images present distinct image styles, it is necessary to from them extract the underlying consistent object descriptors. 
\subsubsection{Motivation behind BFE.}
Convolutional Neural Networks (CNNs) have been widely used as feature extractors in many computer vision tasks \cite{warburg2021bayesian, sarlin2020superglue, godard2019digging} due to their expressive representation power to process image-like data. In a multi-modal detection network, two feature extractors are needed to process thermal and radar images respectively, if other fusion strategies are used other than early (input) fusion. To this end, we adopt the convolution blocks of the YOLOv5s \cite{glenn_jocher_2022_7347926} as our feature extractors due to their small footprint and a proven ability for general object detection tasks. Hereafter we refer to it as the basic Feature Extractor (FE). 

While it is straightforward to use the basic FE for sensor fusion, there are significant limitations that must be addressed when using thermal and mmWave radar sensors. These sensors present unique limitations, such as the low contrast and featureless images of thermal cameras, and the sparse and noisy measurements provided by radar sensors. Intuitively, uncertainty in the extracted features is therefore inevitable as the inputs themselves can be in low fidelity. The challenge here is, however, sorting out all uncertainty sources is impossible and a general uncertainty estimation is needed.
This motivates us to use a Bayesian Neural Network (BNN) \cite{kendall2017uncertainties} in the feature extraction process. BNNs are a sensible choice here because they can model the uncertainty in an agnostic manner by using predictions from weight distributions. We thus propose to upgrade the basic FE to a \underline{B}ayesian \underline{F}eature \underline{E}xtractor (BFE) which offers two advantages: 1) by using a Bayesian feature extractor, we can extract additional information, such as variance, from existing sparse and noisy data, and 2) the sensor uncertainties are mitigated by formulating the feature learning process as an optimization problem with posterior distributions.

\subsubsection{BFE Design Detail}
We now detail the principles of the proposed BFE. In the Bayesian framework, the feature extractor's weights $\mathcal{W} $ are treated as distributions rather than deterministic values. Given the dataset $\mathcal{D}$, the goal is to find the posterior distribution of BFE's weights: 
\begin{equation}
    p(\mathcal{W} \vert \mathcal{D}) = \frac{p(\mathcal{D} \vert \mathcal{W})p(\mathcal{W})}{p(\mathcal{D})}
\end{equation}
However, $p(\mathcal{W} \vert \mathcal{D})$ is intractable since the marginal probability $p(\mathcal{D})$ cannot be evaluated analytically. To solve this issue, Variational Inference (VI) is usually used to approximate the intractable distribution. VI-based methods \cite{kendall2017uncertainties} try to fit the posterior distribution $p(\mathcal{W} \vert \mathcal{D})$ with a tractable parametric distribution, e.g., Gaussian distribution, and then optimize over the parameters of the tractable distributions. 

\begin{figure*}[t]
    \centering
    \includegraphics[width=0.9\linewidth]{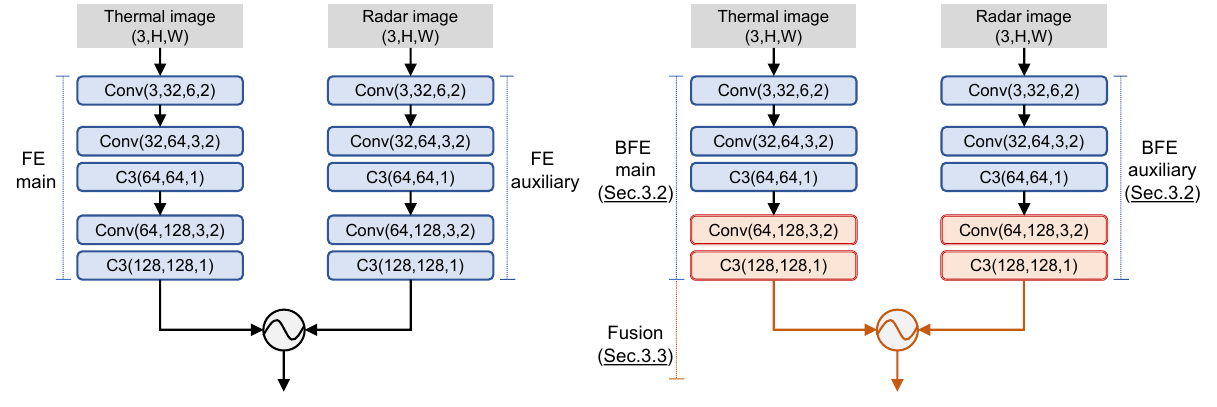}
    \caption{The basic Feature Extractor (FE) and the proposed Bayesian Feature Extractor (BFE): Conv(in,out,kernel\_size,stride): The weights of FE are deterministic, while those of BFE are distributional and approximated with a mixture of two Gaussian distributions (see Sec.4.2 for details).}
    % \chris{@KW, It is better replace this figure with that: left- FE, right - BFE}
    % \Description{The basic Feature Extractor (FE) and the proposed Bayesian Feature Extractor (BFE): Conv(in,out,kernel\_size,stride): The weights of FE are deterministic, while those of BFE are distributional and approximated with a mixture of two Gaussian distributions (see Sec.4.2 for details).}
    \label{fig_feature_extractor}
\end{figure*}

Based on the VI methodology, we adopt Dropout to approximate the posterior distribution. Dropout approximation assumes $p(\mathcal{W} \vert \mathcal{D})$ as a mixture of two Gaussian distributions with small variances and the mean of one is fixed to zero \cite{kendall2017uncertainties}. The original dropout approximation adds a dropout layer to each convolutional layer, and applies dropout during both training and inference phases. In practice, we observed that applying dropout to all convolutional layers decreases the representation ability of a feature extractor. As a result, we only include dropout layers in the final two convolutional blocks in both feature extractors. (markd in red in Figure~\ref{fig_feature_extractor}).

%In sec. we show the performance of different feature extractors. \Kaiwen{add experiments}

As shown in Figure~\ref{fig_feature_extractor}, there are two branches in BFE: \textit{BFE main} and \textit{BFE auxiliary}. BFE main is composed of three convolutional blocks outputting feature maps at $1/2$, $1/4$ and $1/8$ of input size, and BFE auxiliary shares the same architecture as BFE main. Given two cross-modal input data of size $(3,H,W)$ and $(3,H,W)$, the feature maps of the two branches have sizes of $(128, H/8, W/8)$ and $(128, H/8, W/8)$. We input thermal images to the main branch and radar depth images to the auxiliary branch. Nevertheless, the branches are interchangeable, given that they share identical structures. In the next section, we detail how we fuse these cross-modal feature maps.

\begin{figure}[h]
    \centering
    \includegraphics[width=1\linewidth]{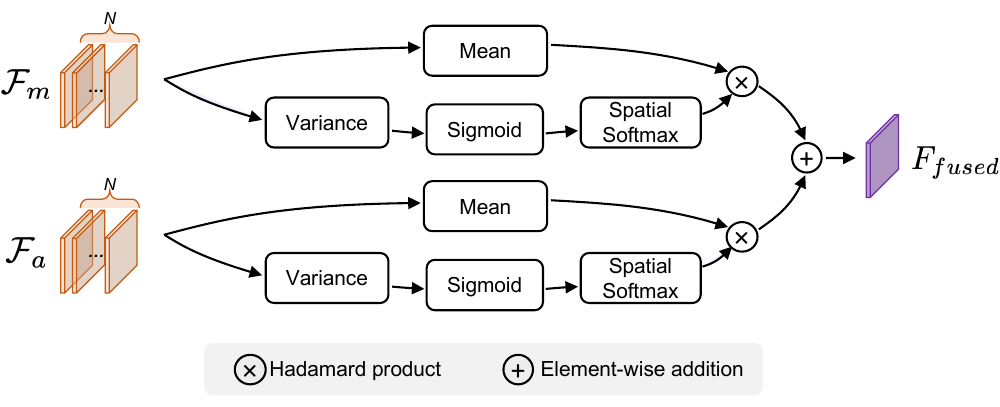}
    \caption{The proposed UGF: $\mathcal{F}_{m}$ denotes the main feature map stack, and $\mathcal{F}_{a}$ the auxiliary feature map stack. 
        % The four sensor fusion strategies.\chris{@KW, we do not need these many fusion strategies but only (the best) one. Choose one and leave others as baselines for comparison.} 
        }
    \label{fig_fusion}
\end{figure}

\subsection{Uncertainty-Guided 
% Feature
Fusion}
\label{sec_ugf}
%sparse cross-modal data make uncertainty a concern for sensor fusion. The data provided by cross-modal sensors could fluctuate significantly  not only present  styles but also show uncertainty as the input information could be insufficient to optimize a deep neural network. \kaiwen{add benefits of Bayesian Neural Network here ..} 
\noindent \textbf{Formulation.} Conventional fusion methods (e.g., \cite{chang2020spatial, abdar2021barf} ignore the importance of feature map uncertainties in guiding multimodal fusion, rendering sub-optimal multimodal detection results. To this end, we propose an \underline{u}ncertainty \underline{g}uided \underline{f}usion module, called UGF, which fuses two feature maps with the guidance of their variance maps provided by the BFE. The principle of the UGF is as follows. To capture the model uncertainty during training, the BFE forward propagates the input data $N$ times and thus produces a stack of $N$ feature maps in each branch, which we refer to as the main feature stack $\mathcal{F}_{m}$ and the auxiliary feature stack $\mathcal{F}_{a}$, which are written as follows:
\begin{equation}
    \begin{aligned}
        \mathcal{F}_{m} = \{F_{m,i}\vert F_{m,i} \in \mathbb{R}^{H\times W \times C}, i=1,2,..,N\}, \\
        \mathcal{F}_{a} = \{F_{a,i}\vert F_{a,i} \in \mathbb{R}^{H\times W \times C}, i=1,2,..,N\}.   \\     
    \end{aligned}
\end{equation}
Figure \ref{fig_fusion} shows the overview of the UGF module. We first derive the mean and the variance map from $\mathcal{F}_{m}$. As the variance indicates where the BFE is uncertain, we apply a sigmoid operation and spatial softmax operation to the variance map, such that the value at each pixel location is converted to the weight of the feature (i.e., the sum of weight is a unit). We apply the same operations to $\mathcal{F}_{a}$. The fused feature map, $F_{fused}$, is the weighted sum of the feature maps of two branches. We put it more formally as
\begin{equation*}
    \begin{aligned}
        F_{fused} & = \phi _{\mu}(\mathcal{F}_{m})\odot \phi _{ss} \phi _{sg} (\phi _{\sigma}(\mathcal{F}_{m})) + \phi _{\mu}(\mathcal{F}_{a}) \odot \phi _{ss} \phi _{sg} (\phi _{\sigma}(\mathcal{F}_{a})),
    \end{aligned}
\end{equation*}
where $\odot$ means the Hadamard product; $\phi _{\mu}$ and $\phi _{\sigma}$  means obtaining the mean and variance of samples, respectively; $ \phi _{ss}$ and $\phi _{sg}$ means applying a sigmoid operation and spatial softmax operation, respectively.  

\begin{figure*}[t]
    \centering  \includegraphics[width=0.95\linewidth]{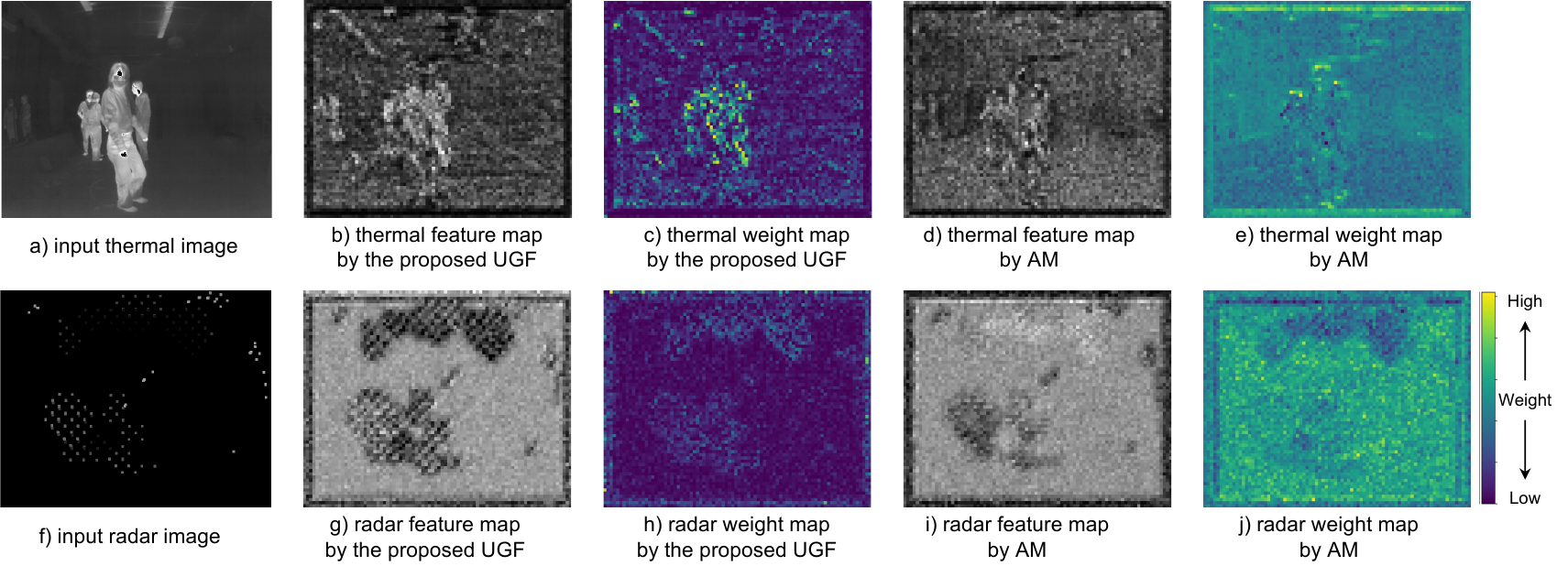}
    \caption{The inputs, feature maps and weight maps of the AM \cite{li2022pedestrian} and the proposed UGF module: AM is focusing more on easily-detected parts, e.g., heads, while UGF is emphasizing on noisy regions, e.g., overlapping human bodies (see highlights in their thermal weight maps). The radar weight maps show that the proposed UGF effectively concentrates on noisy measurements, while AM does not demonstrate a clear concentration.}
    % \Description{The visualization of inputs and features of the Attention Mechanism (AM) and the proposed UGF method.}
    \label{fig_heatmap}
\end{figure*}

\noindent \textbf{Fusion Explanation.} The intuition behind the UGF is that features are not equal to the fused feature map and should be differently treated in the \emph{training phase}. As high variance indicates high uncertainty, an effective training phase should let the model focus more on those features of high uncertainty (i.e., hard examples) so that the model can gradually familiarise itself to handle difficult features coming from both sensors, regardless of what the uncertainty causes are. Through this `hard training', the trained model is expected to predict well even when being fed with high-uncertainty feature maps when it comes to the \textit{inference stage}.

Figure~\ref{fig_heatmap} illustrates the difference between the conventional Attention Mechanism (AM) \cite{li2022pedestrian} and the UGF. It can be observed that for the same input thermal image, the weighted thermal feature map by AM shows that the model is focusing more on regions of human heads, which are typically considered as the characteristics of humans. In contrast, the weighted thermal feature map generated by the UGF emphasizes the overlapping regions of people, where the model typically has difficulty in differentiating between humans and objects. This demonstrates that the UGF forces the model to concentrate on challenging regions during the training in order to improve the inference robustness of the trained model. 

\subsection{Multiscale Detection Net}
\label{sec_mdn}
\begin{figure*}[h]
    \centering
    \includegraphics[width=0.9\linewidth]{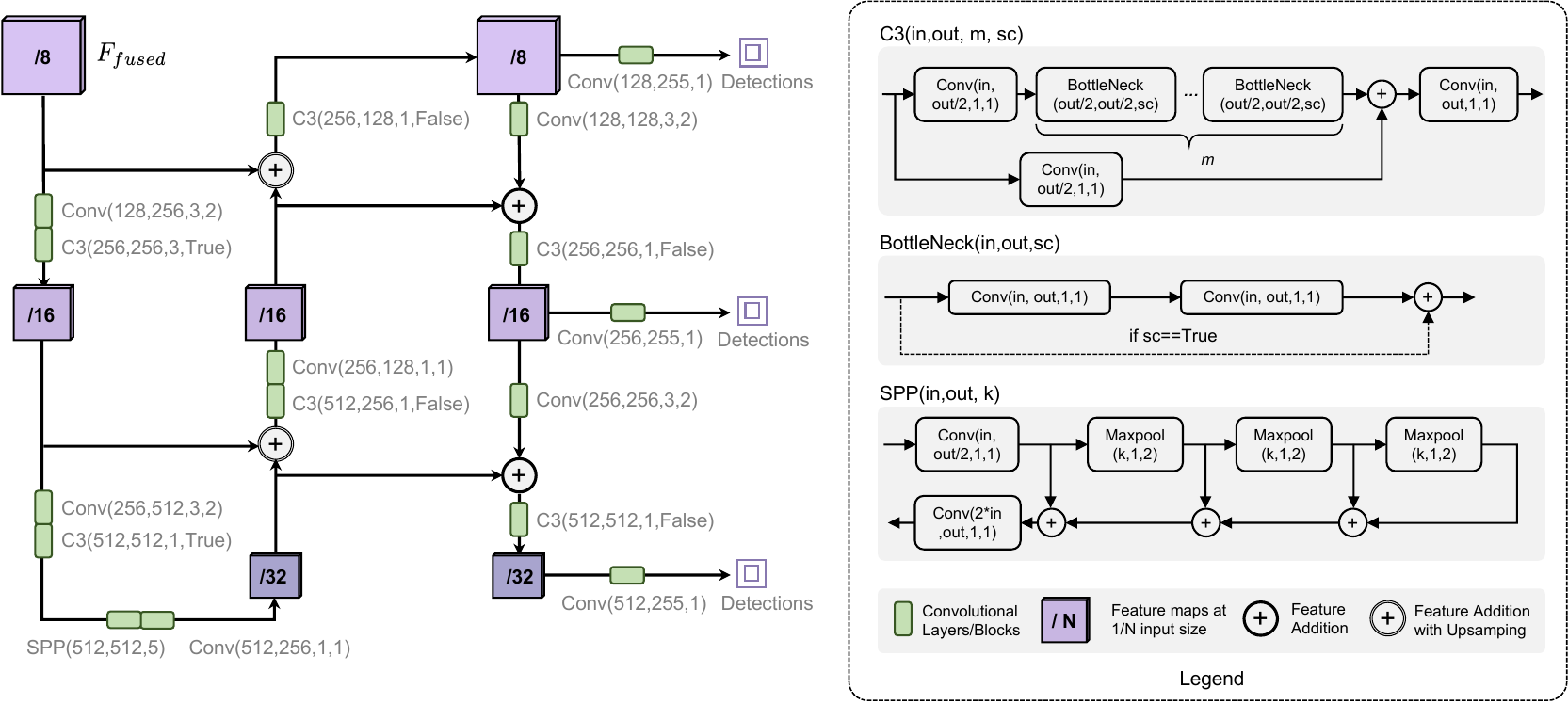}
    \caption{The Multiscale Detection Net: Feature maps are extracted and processed at three scales, $1/8$, $1/16$ and $1/32$ of the input image size. Detections are generated at each scale and then summarized.}
    % \Description{The Multiscale Detection Net.}
    \label{fig_mdn}
\end{figure*}

% uncertainty aware thermal and mmWave radar fusion UTM-Fusion
Once the fused feature map is obtained, it will be passed to the Multiscale Detection Net (MDN), which is based on the YOLOv5s \cite{glenn_jocher_2022_7347926} network. The YOLOv5s network is explained in detail in reference \cite{glenn_jocher_2022_7347926}, but the principle of MDN is also provided in this section for completeness. Figure~\ref{fig_mdn} shows the network architecture of MDN, which is composed of Convolutional layers or blocks, including the single convolutional layer, the C3 block and the Spatial Pyramid Pooling (SPP) block. The convolutional layers and blocks progressively extract feature maps at $1/8$, $1/16$ and $1/32$ of the input image size. The detection is made on feature maps at three scales, which accounts for small, medium, and large objects. Finally, the Non-Maximum Suppression (NMS) method is used to generate the final detection results.
% We kindly refer interested readers to [] for detailed descriptions of the YOLOv5s architectures.

% \kaiwen{@Chris, to what extend should we elaborate the networks of YOLOv5s?} \chris{as much as possible. Presented it as a new contribution is desirable. The current methodology is a bit thin}.

%% file: sections/5_experiments.tex
\section{Evaluation}

% \kaiwen{4.1 has not been revised by Kaiwen yet.}
% \kaiwen{4.2, 4.3, 4.4 are ready for Chris to revise.} \chris{KW, revise 4.1 in a way to reflect our contribution to the dataset. This includes the 'duration' of the entire data collection phase (3 months?) How many volunteers participated in the experiment? You can have a look at the milliEye work first and ask QY any missing information regarding it.}

\subsection{Hardware and Data Collection}
As there is currently a lack of publicly available datasets that incorporate both thermal camera and mmWave radar sensors, we designed a sensing platform and gathered data from a laboratory testbed for analysis. We are committed to making this dataset available to the research community and plan to release it publicly upon acceptance of our paper.

\begin{figure}[!h]
    \includegraphics[width=\linewidth]{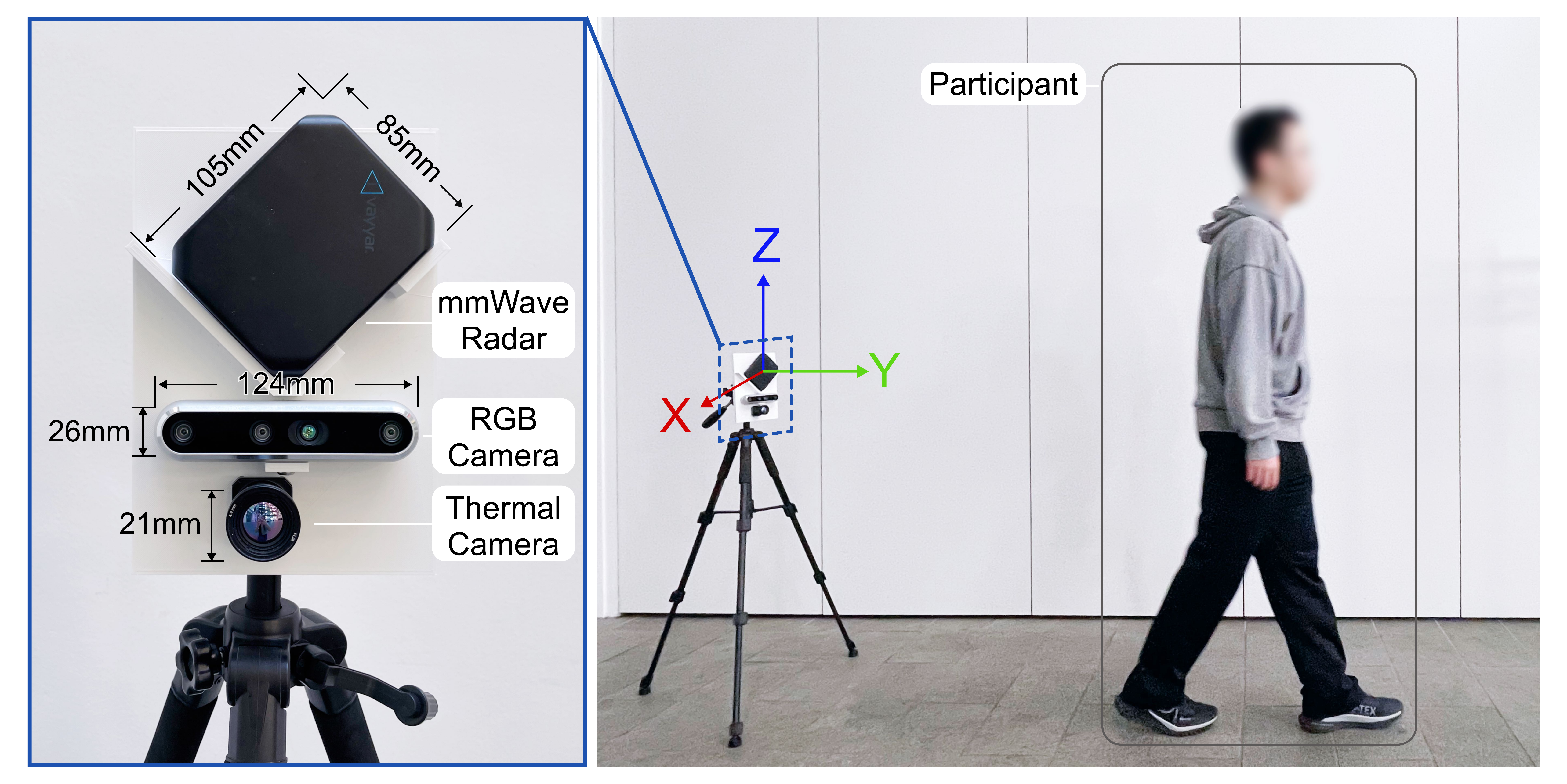}
        % \captionsetup{width=0.9\textwidth}
    \caption{\label{fig:sensing_platform} The sensing platform setup. 
    % \chris{@QY, please use the image where the sensor is attached to the wall.}\kaiwen{might be better with a human in the scene.}
    Note that the RGB camera is only used for the pseudo-labelling in the training stage and is NOT used in the inference stage.} 
\end{figure}

\noindent \textbf{Sensing Platform Setup.}
Figure~\ref{fig:sensing_platform} presents the sensing platform comprising a mmWave radar, a thermal camera, and an RGB camera, affixed to a 3D printed carrier board. 
% and can be flexibly deployed in various environments. 

The mmWave radar, Vayyar vTrigB \cite{vayyar_2023}, is the current best-in-class 4D radar operating between 62-69GHz. 
% The carrier frequency is set up as 62-69GHz.
It has 20 TX and 20 RX antennas, capable of producing IQ signals with point cloud data at a density of 1000 to 3000 points per frame. The weight of this probe is 110g and the probe size is 105mm × 85mm. The mmWave radar is capturing at a frequence of $6 \sim 13$  Hz. 
% The thermal camera is the FLIR Boson 640 \cite{flir_boson640}, which has a resolution of 640 × 512 and an HFov of 95°. This thermal camera works with a frame rate of 9 Hz and the scene temperature range can detect up to 140 °C with high gain and 500 °C with low gain. The weight of this probe is 79g and the probe size is 21mm × 21mm × 11mm. The RGB camera used for labelling is Intel RealSense D455 \cite{intel_realsense}. The RGB images are captured with a resolution of 640 × 480 and the Fov of 95 × 65°. The weight of this probe is 389g and the probe size is 124mm × 26mm × 29mm.

The thermal camera, the FLIR Boson 640 \cite{flir_boson640} thermal camera, has a 640 × 512 resolution, 9 Hz frame rate, and a 95° HFov that can detect temperatures up to 140 °C with high gain and 500 °C with low gain. This 79g probe has a small size of 21mm × 21mm × 11mm. 

The Intel RealSense D455 \cite{intel_realsense} RGB camera has a working frequency of up to 90 Hz, a resolution of 640 × 480, and a 95 × 65° Fov. It measures 124mm × 26mm × 29mm and weighs 389g. We employ the RGB camera solely for labeling purposes.

\begin{figure*}[!t]
    \centering
    \begin{subfigure}[c]{.19\textwidth}
        \includegraphics[width=\linewidth]{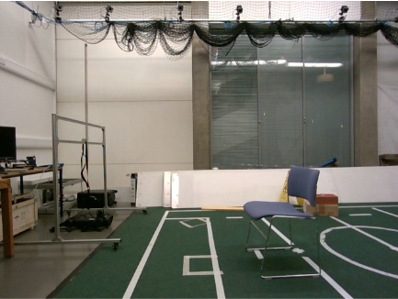}
        \caption{\label{fig:env_office1} Office 1} 
    \end{subfigure}
    \begin{subfigure}[c]{.19\textwidth}
        \includegraphics[width=\linewidth]{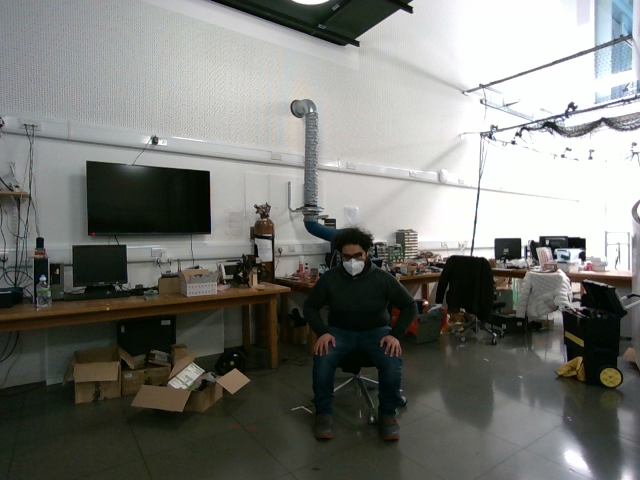}
        \caption{\label{fig:env_office2} Office 2 } 
    \end{subfigure}
    \begin{subfigure}[c]{.19\textwidth}
        \includegraphics[width=\linewidth]{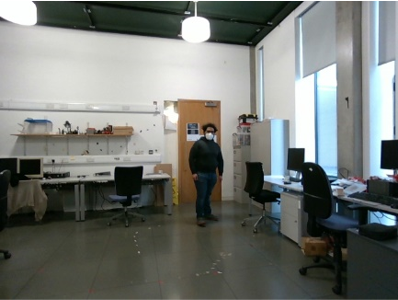}
        \caption{\label{fig:env_office3} Office 3} 
    \end{subfigure}
    \begin{subfigure}[c]{.19\textwidth}
        \includegraphics[width=\linewidth]{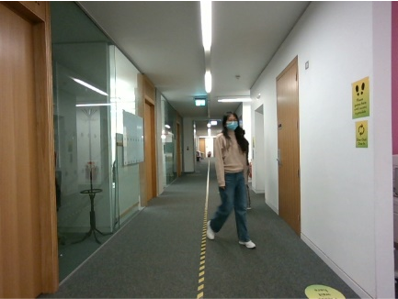}
        \caption{\label{fig:env_corridor1} Corridor 1} 
    \end{subfigure}
    \begin{subfigure}[c]{.19\textwidth}
        \includegraphics[width=\linewidth]{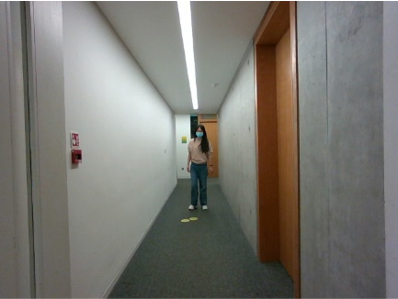}
        \caption{\label{fig:env_corridor2} Corridor 2} 
    \end{subfigure}

    \begin{subfigure}[c]{.19\textwidth}
        \includegraphics[width=\linewidth]{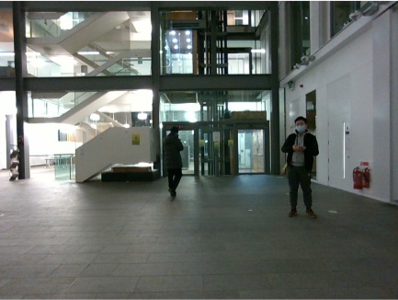}
        \caption{\label{fig:env_atrium1} Atrium 1} 
    \end{subfigure}
    \begin{subfigure}[c]{.19\textwidth}
        \includegraphics[width=\linewidth]{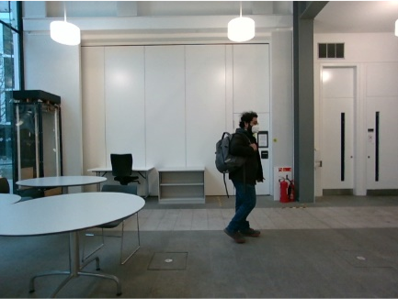}
        \caption{\label{fig:env_atrium2} Atrium 2} 
    \end{subfigure}
    \begin{subfigure}[c]{.19\textwidth}
        \includegraphics[width=\linewidth]{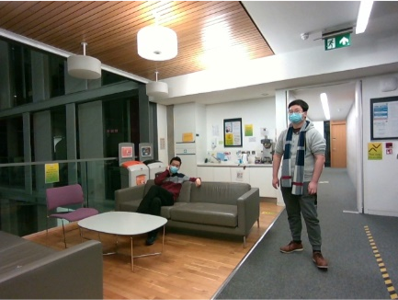}
        \caption{\label{fig:env_kitchen} Kitchen} 
    \end{subfigure}
    \begin{subfigure}[c]{.19\textwidth}
        \includegraphics[width=\linewidth]{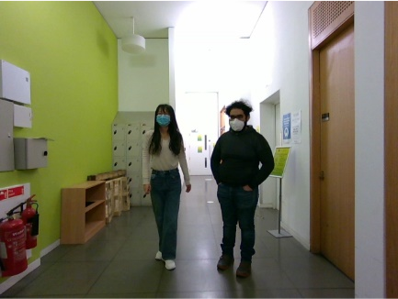}
        \caption{\label{fig:env_corridor3} Corridor 3} 
    \end{subfigure}
    \begin{subfigure}[c]{.19\textwidth}
        \includegraphics[width=\linewidth]{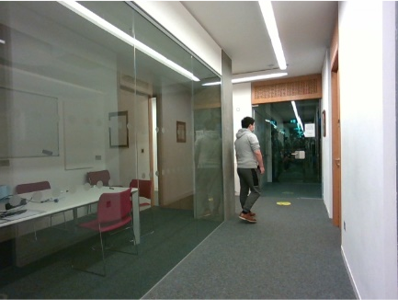}
        \caption{\label{fig:env_corridor4} Corridor 4} 
    \end{subfigure}
    \caption{\label{fig_our_dataset} Ten scenes with different illuminations, spatial configurations, and temperature conditions.}
\end{figure*}

\noindent \textbf{Data Collection.} 
To ensure the dataset's versatility, we collected human detection data\footnote{This work has received the ethical approval XXX (no name for now due to double-blind review).} from various indoor environments, including offices, corridors, atriums, and kitchens. These settings possess varying lighting, spatial arrangements, and temperature conditions, as illustrated in Figure~\ref{fig_our_dataset}.
% During the experiment, volunteers randomly walk, stand, sit, or lie down in front of the sensors. Figure~\ref{fig_our_dataset} shows the lab testbed data collection environments. 
% we collect the data in 10 different places including office rooms, corridors, atriums, and the kitchen. our dataset is collected using the infrastructure sensing platform. 
% We invited 4 volunteers to walk around with varying poses in the scene while all our platform was sensing. We finally gathered 24,000 frames in total, each frame contains 1 to 4 people. We split the dataset into a 15k training set, a 4k validation set and a 4k test set.
We recruited volunteers to move around in the scene with different poses, while our platform detected and recorded the activity. Ultimately, we collected a total of 24,241 frames with each frame featuring 1 to 4 individuals and 12 volunteers participating in our data collection to ensure diversity. 
We randomly separated the frames into three distinct sets: 15,641 for training, 4,300 for validation, and 4,300 for testing.

To speed up the labelling process, we utilized the advanced vision-based detection tool detectron2 \cite{detectron2}. We initially obtained precise ground truth boxes on the RGB images. By leveraging the known intrinsic and extrinsic parameters of the RGB camera, thermal camera, and mmWave radar, we are able to effortlessly derive the ground truth boxes on both thermal and radar depth images through pose transformation.

% \begin{figure}[h]
%     \centering
%     \includegraphics[width=0.9\linewidth]{figures/firefighter_drill.pdf}
%     \caption{\label{fig:dataset_SFRS} The firefighter drill dataset collection setup.}
% \end{figure}

\subsection{Implementation Details}
The previously mentioned components - BFE, UGF, and MDN - make up the complete pipeline of \sysname, which we train in an end-to-end manner. 
To begin with, the thermal images and radar depth images are synchronized using the Robot Operating System \cite{quigley2009ros} and then resized to a resolution of $640 \times 512$.
In the UGF module, we set the dropout rate to $p=0.2$ and the number of forward passes to $N=5$. 
% based on the results of the testflight \chris{KW, what do you mean 'testflight'?} experiment (refer to Section \ref{sec_bfe_approximate}). 
To optimize our model, we utilize a stochastic gradient descent optimizer with an initial learning rate of $0.01$, a momentum of $0.937$, and a weight decay of $5 \times 10^{-4}$. We decrease the learning rate linearly over time until it reaches zero at the final epoch. Our training process spans 100 epochs, as we observe that the model's performance plateaus after this point. The selection of the best model is based on its performance on the validation set. 
% \chris{KW, remember to specify how we split the dataset into train, val, and test. Make sure we give a number of samples per set}\kaiwen{QY add dataset split}.
% The batch size for training is 8. 
% We implement UGF on the PyTorch \cite{paszke2019pytorch} platform. The weights in UGF are random initialized and trained on ... GPUs with batch size of 8. Dropout is placed after layer 3 and 4 in training stage and the dropout rate is set to be 0.2. In model training, the input thermal images and radar depth maps are resized to be $640\times640$ for our dataset and $160\times160$ for the firefighter drill dataset.

% need description about learning rate, warmup scheme, ...
% \qiyue{@KAIWEN, could you add more implementation details here?}

\subsection{Evaluation Metrics}
We evaluate the performance of \sysname using predominant object detection metrics, including \textbf{precision}, \textbf{recall}, \textbf{max F1 score (mF1)} and \textbf{Average Precision (AP)} \cite{pascalvoc2007} at different Intersection-over-Union (IoU) thresholds. The IoU is defined as the ratio of the intersection between predicted and ground truth bounding boxes to the union of the same:
\begin{equation}\label{eq:iou}
    IoU=\frac{Predicted Box \cap GT Box}{Predicted Box \cup  GT Box},
\end{equation}
and a detection is considered to be a true positive if the IoU is greater than a predefined IoU threshold. When assessing all detections, the precision and recall values are calculated as: 
\begin{equation}
    \begin{aligned}
    Precision&=\frac{True Positives}{True Positives+False Positives}, \\
    Recall&=\frac{True Positives}{True Positives+False Negatives},
    \end{aligned}
\end{equation}
% where TP, FP, FN are true-positives, false-positives and false-negatives respectively. 

AP is the average detection precision under different recall values, i.e., the area under the PR curve. We follow \cite{lin2014microsoftcoco} and evaluate AP with IoU thresholds from $0.50$ to $0.95$, denoted as AP$_{50}$, AP$_{65}$, .., AP$_{95}$, and calculate the mean AP over all IoU thresholds, denoted as AP$_{50:95}$. 

F1 score is calculated as 
\begin{equation}
F1 = 2 \times \frac{Precision \times Recall}{Precision + Recall}
\end{equation}
and the mF1 is the maximum F1 overall precision and recall values. Similar to AP and for fairness reasons, we evalaute the mF1 with different IoU thresholds from $0.50$ to $0.95$, denoted as mF1$_{50}$, mF1$_{65}$, ..,mF1$_{95}$, and calculate the mean max F1 score over all IoU thresholds, denoted as mF1$_{50:95}$.

% mAP is the mean of per-class AP, and it is equivalent to AP in our human detection setting since there is only one class - human. To calculate AP, the confidence threshold is varied from 0 to 100\% and precision and recall are calculated at each point, which forms a P-R curve. The AP is then the area below the obtained P-R curve. 

% The mAP with different IoU threshold is used in evaluating our model, denoted as mAP@0.50, mAP@0.65,mAP@0.75, mAP@0.85 and mAP@50:95 (averaged mAP over IoU values from 0.50 to 0.95). 

% \subsubsection{mean Average Precision (mAP)}

\subsection{Experimental Results}
The following sections give a numerical analysis and a video demo shot in the dark environment can be found at \url{https://github.com/ramdrop/utm}.
% \chris{KW, QY, please add a youtube link here when 'the human detection in the dark environments' is done.} for the ease of visualization.

\subsubsection{Competing methods}
\label{sec_competing_methods}
We compare the proposed \sysname against the following methods:

% \begin{figure*}
%     \centering
%     \includegraphics[width=5.5in]{figures/fusion_baselines.pdf}
%     \caption{The fusion baselines. \chris{can be deleted if non-needed.}}
%     % \Description{The fusion baselines.}
%     \label{fig_fusion_baselines}
% \end{figure*}
\textbf{SOD} \cite{liu2019salient} fuses the RGB channel and the depth channel by composing a four-channel image. We follow this idea to fuse the single-channel thermal image and radar depth image. Specifically, we concatenate two thermal images and one radar depth map to form a pseudo RGB image. To ensure fair comparison, we retain the other architectures intact while replacing our dual-branch BFE with a single-branch FE.

\textbf{TOD} \cite{Kristo2020Thermal} is a thermal camera-based object detection method, which we evalaute by replacing our dual-branch BFE with a single-branch FE while maintaining the same architectures. Furthermore, in order to examine the impact of mmWave radar sensor, we replace the input thermal image with the radar depth image, which we term as \textbf{TOD-Radar}.

\textbf{MilliEye} \cite{shuai2021MilliEye} uses mmWave radar and RGB cameras for human detection, where the fusion is performed at the decision level. It consists of an RGB image-based object detector trained on the public datasets COCO \cite{lin2014microsoftcoco} and ExDark \cite{loh2019exdark}, a radar-based object tracker that produces box proposals to combine with image-based proposals, and an ROI-wise refinement head. We fine-tune MilliEye on our dataset.
% \chris{KW, again, don't call it `lab testbed dataset" just say it our dataset'}

\textbf{Vanilla Addition (VA)} \cite{chang2020spatial} is a fusion technique that operates at the feature level. To ensure fairness, we evaluate it by solely substituting the proposed UGF module of \sysname with VA. Specifically, the BFE propagates the input data $N$ times, resulting a stack of $N$ feature maps at each branch during the training. We obtain a single average feature map by computing the mean of the stack of feature maps. The two mean feature maps are added together and passed to the subsequent layers. 
% VA is trained with the same parameters as the \sysname model.

\textbf{Attention Mechanism (AM)} \cite{li2022pedestrian} is another feature-level fusion method. We evaluate it by solely substituting the proposed UGF module of \sysname with AM. we first obtained mean feature maps from each branch, then convolved and applied a softmax operation on the radar branch to generate an attention mask. In parallel, we applied convolutional operations on the thermal branch to generate a query map, which was multiplied by the attention mask. The resulting feature map was then passed to the subsequent layers. 
% Training of AM is done using the same set of parameters as used for the \sysname model.

% \chris{Any more baselines about the fusion methods?}\kaiwen{That's all.}
% \kaiwen{vanilla add, attention to be added.}

\subsubsection{Overall Performance}

\begin{figure*}[ht]
    % \begin{minipage}[t]{0.5\linewidth}
      \centering
      % \captionsetup{width=0.9\linewidth}
      \includegraphics[width=\linewidth]{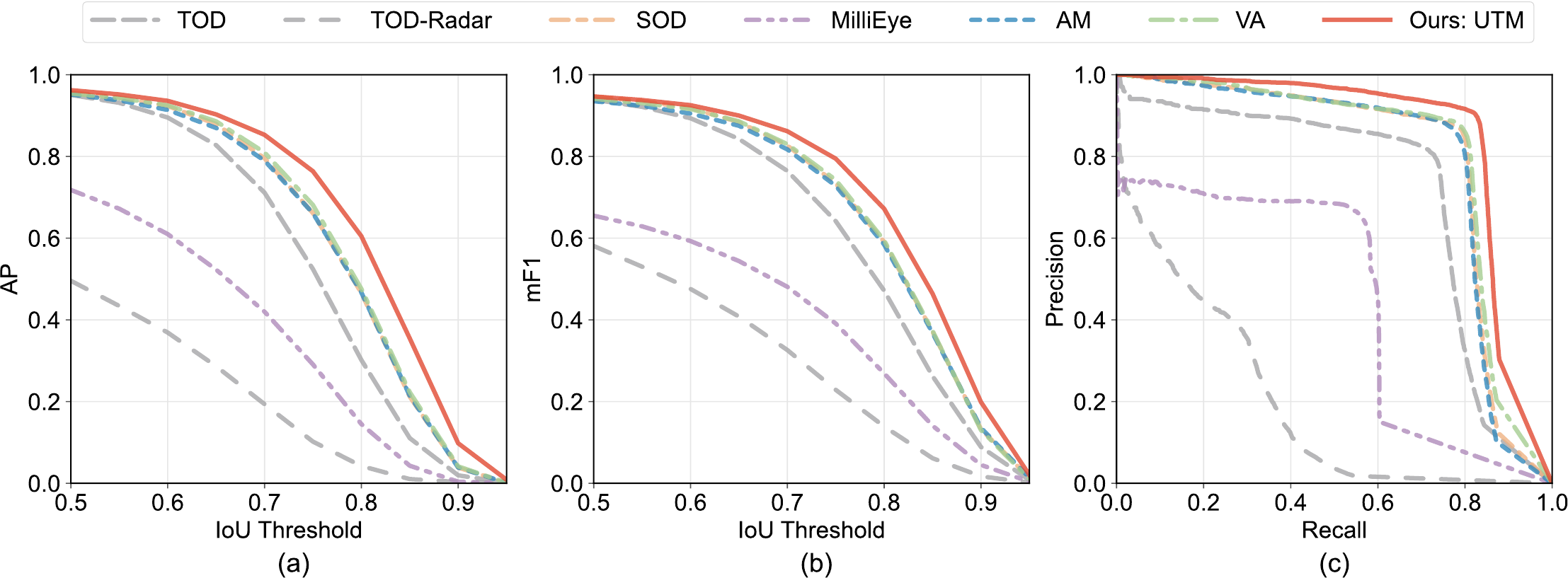}
      \caption{The mAP, mF1 and precision-recall curve of different methods under different IoU thresholds with a NMS IoU threshold of $0.6$ and precision and recall are with a detection IoU threshold of $0.7$. Our \sysname outperforms the competing methods by a large margin in the three metrics.}
      \label{fig_map}
    % \end{minipage}%
  \end{figure*}

\begin{table*}[ht]
    \centering
    \caption{ The AP and mF1 of different methods under different IoU thresholds with an NMS IoU threshold of 0.6. }
    \scalebox{0.89}{\input{./tables/result_indoor_map.tex}}
    % \scalebox{1}{\input{./tables/result_indoor_map.tex}    }
    % \chris{@KW, remember to add the reference to the methods in the table. The two methods w.o. can have no reference need to be discussed in our meeting.}
    \label{tab_lab_map}
\end{table*}

Table~\ref{tab_lab_map} shows the AP and mF1 of various methods under different IoU thresholds when evaluated on our dataset with an NMS IoU 
% \chris{KW, did you spell NMS and IoU in previous sections? If not and this is the first usage, please spell it now}
threshold of 0.6 and Figure~\ref{fig_map} illustrates the AP, mF1 and precision-recall curve of different methods on our dataset. We note that our evaluation involves three levels of fusion: input fusion, feature fusion, and decision fusion. SOD represents the input (early) fusion technique. Regarding feature fusion, we tested three methods, namely AM, VA, and \sysname. And MilliEye is a decision (late) fusion-based approach.

% Based on the evaluation results represented in Figure~\ref{fig_map}, it is evident that MilliEye performs the worst among the aforementioned methods, as its mAP, mF1 and precision-recall curve are the lowest. 
Among sensor fusion methods, MilliEye performs the worst with the lowest mAP$_{50:95}$ of $0.343$ and mmF1$_{50:95}$ of $0.375$ and its curve is clearly surpassed by those of other methods in Figure~\ref{fig_map}. The under performance of MilliEye can primarily be attributed to the fact that its image-based object detector is trained on public RGB object detection datasets while our dataset comprises thermal and radar-based imagery, leading to a domain gap. Consequently, the performance of the MilliEye system is limited even after fine-tuning its image-based object detector using thermal images and mmWave radar point clouds.

SOD, AM, and VA exhibit similar values for AP and mF1 across various IoU thresholds in Table~\ref{tab_lab_map}, as also demonstrated by their overlapping curves in Figure~\ref{fig_map}. This suggests that input fusion and feature fusion techniques perform comparably. Nevertheless, \sysname surpasses them by a substantial margin, with a mAP$_{50:95}$ of $0.644$ and a mmF1$_{50:95}$ of $0.672$, outperforming SOD, AM and VA with mAP$_{50:95}$ boosts of $9.5\%$, $10.2\%$ and $8.4\%$, and mmF1$_{50:95}$ boosts of $6.8\%$, $5.8\%$ and  $5.8\%$, respectively.

Although both AM and \sysname are attempting to shift the model's focus to particular regions during training, the findings demonstrate that \sysname, leveraging the proposed UGF, is superior to AM in integrating thermal images and radar depth images. Compared with VA, the UGF of \sysname is able to learn the optimal fusion weights for each sensor, which is more flexible than the fixed weights in VA. Note that as illustrated in Figure~\ref{fig_map}, the performance gaps between \sysname and other competing methods are more significant for high IoU thresholds (IoU = $0.7 \sim 0.9$). Successful detection under a high threshold means the model producing more precise bounding boxes that accurately match the ground truth box. Detections under high IoU thresholds are particularly desired in scenarios where safety is crucial. Therefore, \sysname is superior in safety-related scenarios.

\begin{figure*}[ht]
    \centering
    \begin{subfigure}{0.47\textwidth}
        \includegraphics[width=\linewidth]{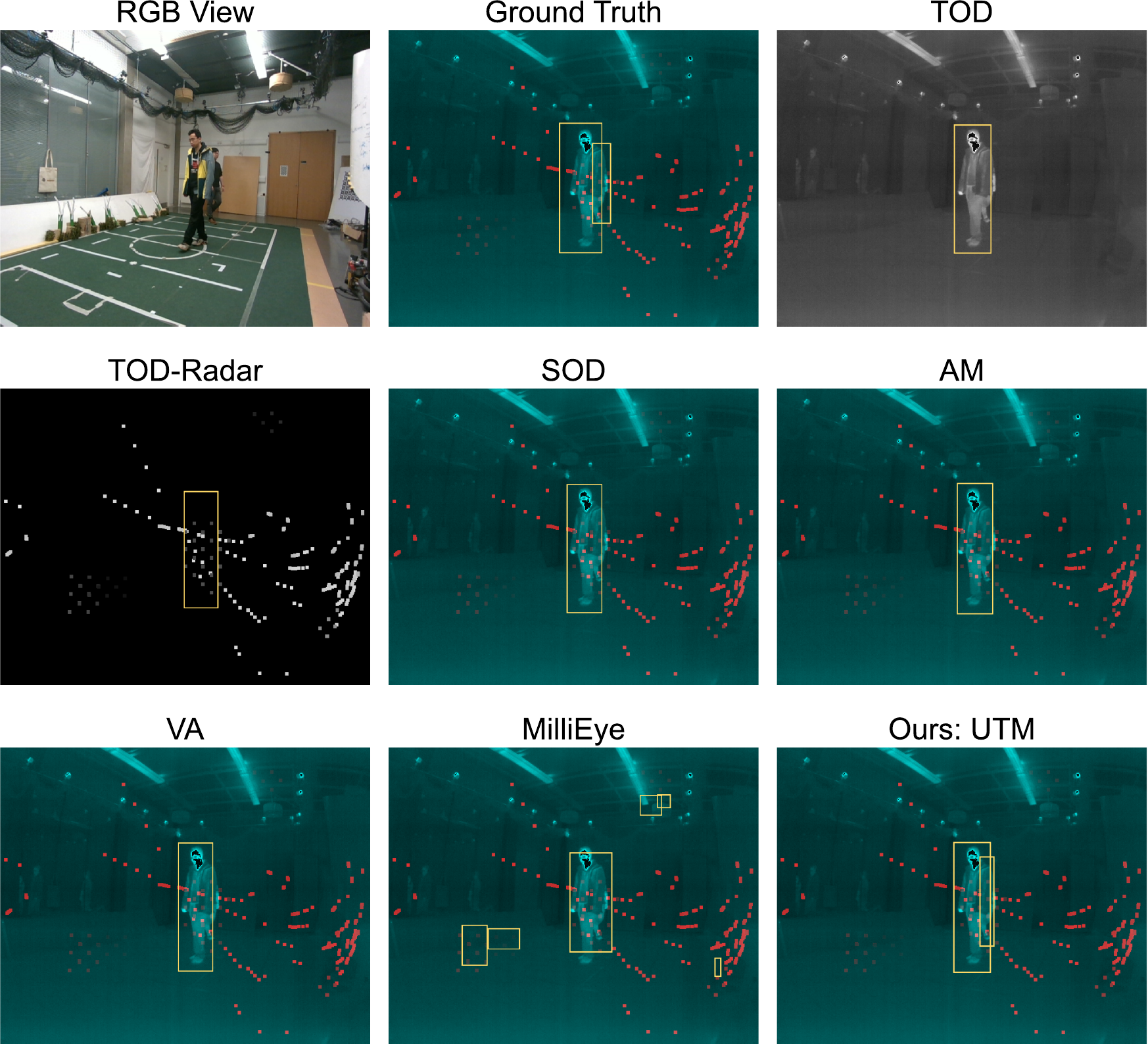}
        \caption{Scene 1}
        \label{fig_qua_lab_1}
    \end{subfigure}
    \hfill
    \begin{subfigure}{0.47\textwidth}
        \includegraphics[width=\linewidth]{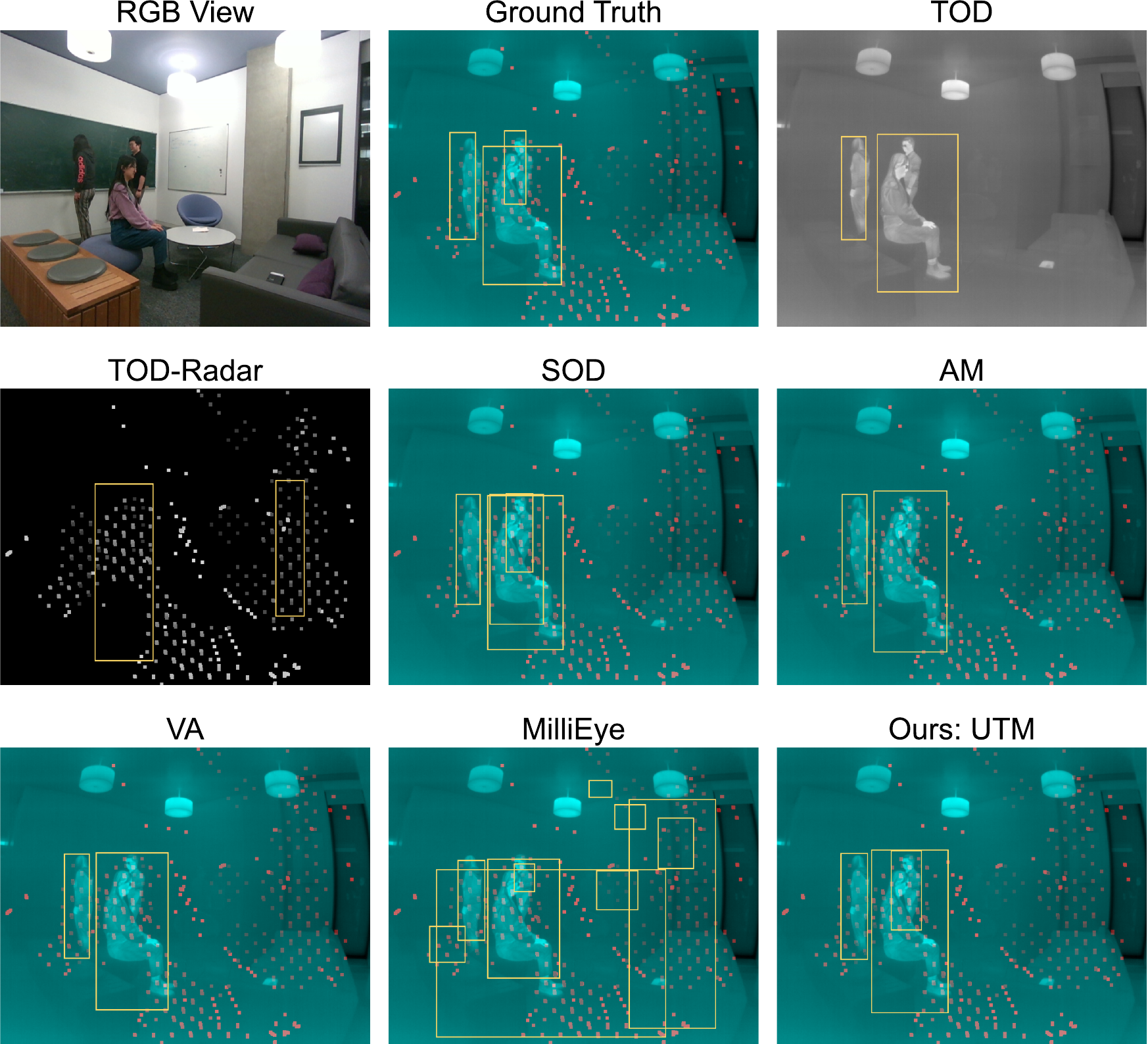}
        \caption{Scene 2}
        \label{fig_qua_lab_2}
    \end{subfigure}
    \caption{The qualitative human detection results on our dataset: it is shown that detecting all human bodies when they are overlapping can be difficult. \sysname was able to successfully detect all human bodies in the scene, while other methods failed. (To visualize the thermal image and radar point cloud together in the same image, we create a pseudo RGB image by assigning the red and green channels to the thermal image and the blue channel to the radar point cloud.)}
    \label{fig_qua_lab}
\end{figure*}

Figure \ref{fig_qua_lab} presents the qualitative results of different methods on our dataset. It is shown that detecting all human bodies when they are overlapping can be difficult. However, UTM was able to successfully detect all human bodies in the scene, while other methods failed.

\subsubsection{Impact of Sensors}
\label{sec_sensor}
The superiority of SOD over TOD and TOD-Radar is expected, as evidenced by the mAP$_{50:95}$ values of $0.588$, $0.527$, and $0.194$, respectively. This is due to the fact that more sensor information leads to better detection performance. Thermal images significantly improve the performance of the SOD, more than radar images. This is due to the fact that radar point clouds tend to be sparser and noisier than thermal images, providing less useful information for object detection. This trend is also evident in Figure~\ref{fig_map}, where the SOD method betters the TOD and TOD-Radar by large gaps.

\subsubsection{Impact of IoU threshold for NMS}
\label{sec_nms}
\begin{figure}[ht]
    % \begin{minipage}[t]{0.5\linewidth}
      \centering
      % \captionsetup{width=0.9\linewidth}
      \includegraphics[width=\linewidth]{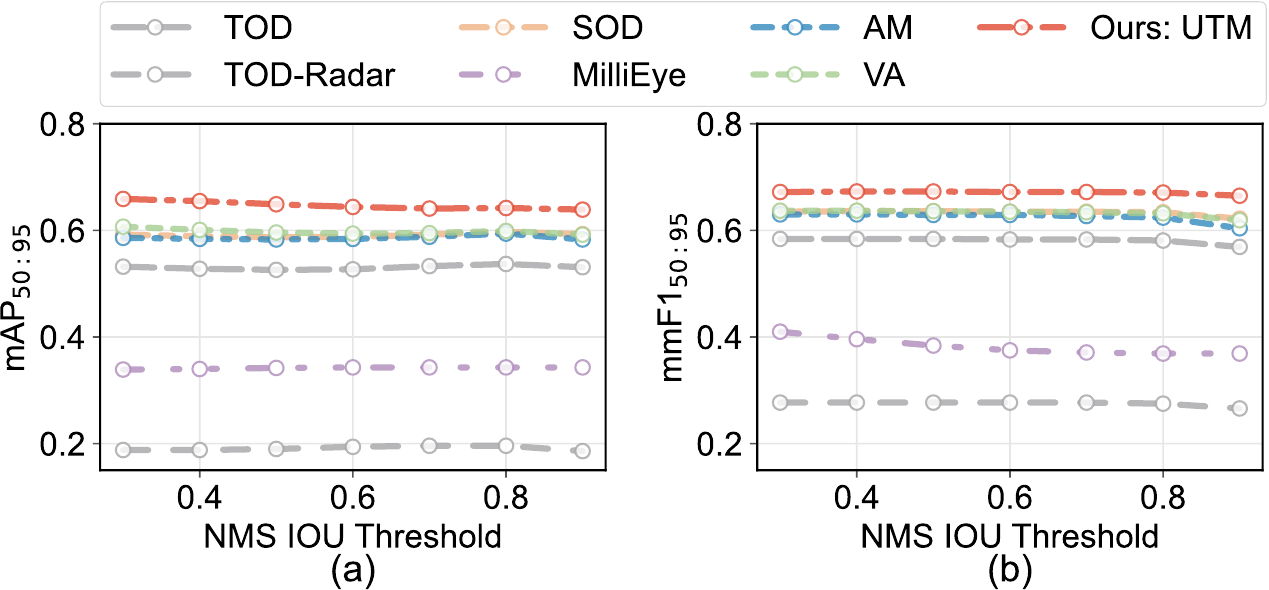}
      \caption{The mAP and mean Max F1 of different methods under different NMS IoU thresholds. Our \sysname consistently performs well across varying NMS IoU thresholds.}
      \label{fig_nms_threshold}
    % \end{minipage}%
  \end{figure}

NMS is a widely-used technique in object detection that helps to eliminate overlapping bounding boxes and is also used in \sysname. The IoU threshold used in NMS can significantly impact the precision and recall values of an object detection model. For this consideration, we evaluate the sensitivity of our \sysname to the IoU threshold for NMS. Figure~\ref{fig_nms_threshold} presents the mAP$_{50:95}$ and mean max F1$_{50:95}$. It can be seen that our \sysname shows a negligible performance change as the IoU threshold for NMS increases from 0.30 to 0.90 while maintaining the best mAP$_{50:95}$ and mean max F1$_{50:95}$ among all methods. This implies satisfying model robustness against this important hyperparameter. 

\subsubsection{Impact of BFE Parameters}
\label{sec_bfe_approximate}
% \usepackage{multirow}
% As previously described in \ref{sec_bfe}, MC Dropout is used to approximate a BNN. Here, we intexplore various approximations of a Bayesian feature extractor by applying dropout at distinct layers with various dropout rates. 

\begin{figure}[ht]
    % \begin{minipage}[t]{0.5\linewidth}
        \centering
        % \captionsetup{width=0.9\linewidth}
        \includegraphics[width=\linewidth]{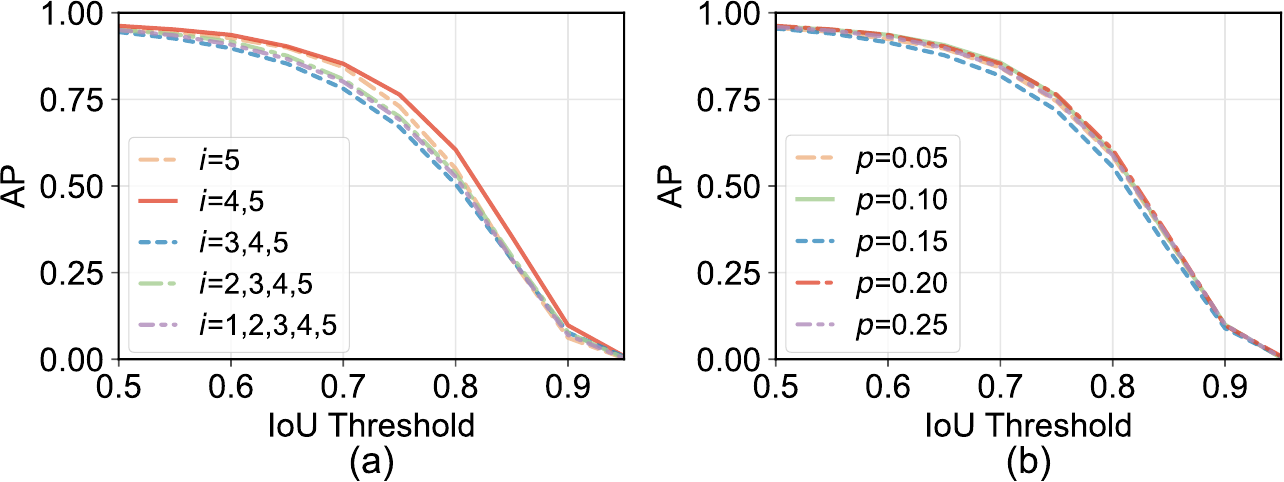}
        \caption{The mAP of BFE with varying configurations. 
        % \chris{This figure requires a zoom-in operation. @KW, let's discuss it in the meeting.}\kaiwen{remove it if necessary}
        % \kaiwen{@Chris, the line width should be kept as it is now, because a larger line width would make the gaps harder to visualize.}
        }
        \label{fig_map_abalation}
    % \end{minipage}%
\end{figure}

Table~\ref{table_bfe_changing_layer} and Figure~\ref{fig_map_abalation}.a) display the AP of BFE with dropout enabled at various layers when $p$ is fixed at $0.20$. It can be seen that the highest mAP$_{50:95}$ was attained by enabling dropout at the last two layers. We hypothesize that this is because insufficient dropout layers fail to adequately approximate a BNN, while excessive dropout layers may compromise the feature extractor's representational capacity.

Table~\ref{table_bfe_changing_p} and Figure~\ref{fig_map_abalation}.b) present the AP of BFE with varying values of $p$ while dropout is enabled in $i=4,5$ layers. It can be observed that \sysname achieves the highest highest mAP$_{50:95}$ when $p=0.20$. However, it is worth noting that the AP gaps among the different BFEs are relatively small, suggesting that \sysname's performance is not significantly affected by changes in the dropout rate.

\begin{table}
    \centering
    \caption{The mAP of BFE with dropout enabled at various layers when p=0.20.}
    \label{table_bfe_changing_layer}
    \begin{tabular}{l|ccccc|c} 
    \hline
            & \multicolumn{5}{c|}{Dropout @ $i^{th}$ layer}                            & \multirow{2}{*}{mAP$_{50:95}$ ↑}  \\
            & 1            & 2            & 3            & 4            & 5            &                                 \\ 
    \hline
    $i=5$       & \textbf{}    &              &              &              & $\checkmark$ & 0.619                           \\
    $i=4,5$    & \textbf{}    &              &              & $\checkmark$ & $\checkmark$ & \textbf{0.644}                  \\
    $i=3,4,5$   &              &              & $\checkmark$ & $\checkmark$ & $\checkmark$ & 0.595                           \\
    $i=2,3,4,5$  &              & $\checkmark$ & $\checkmark$ & $\checkmark$ & $\checkmark$ & 0.611                           \\
    $i=1,2,3,4,5$ & $\checkmark$ & $\checkmark$ & $\checkmark$ & $\checkmark$ & $\checkmark$ & 0.604                           \\
    \hline
    \end{tabular}
\end{table}

\begin{table}
    \centering
    \caption{The AP of BFE with varying values of $p$ while dropout is enabled in $i=4,5$ layers.}
    \label{table_bfe_changing_p}
    \scalebox{0.9}{
    \begin{tabular}{l|lllll} 
    \hline
                    & p=0.05                    & p=0.10                    & p=0.15                    & p=0.20                    & p=0.25                     \\ 
    \hline
    mAP$_{50:95}$ ↑ & \multicolumn{1}{c}{0.634} & \multicolumn{1}{c}{0.642} & \multicolumn{1}{c}{0.619} & \multicolumn{1}{c}{\textbf{0.644}} & \multicolumn{1}{c}{0.637} \\
    \hline
    \end{tabular}}
\end{table}

% \subsection{Robustness in Challenging Scenarios}
% Thick smoke is a common event that occurs in many emergency incidents such as firefighting. In this experiment we examine the robustness and potential use of our human detection module in smoke-filled environments. We test our UGF system on our firefighter drill dataset, the mAP results are shown in Table \ref{tab:drill_map}. The firefighter drill dataset contains the thermal images and mmWave radar point clouds come from lower-cost devices compared with our dataset, which is more challenging both on the perception sources and test scenes. Our UGF shows mAP of 0.428 under IoU 0.75 and average mAP of 0.404. 

% \begin{table}[ht]
%     \centering
%     % \scalebox{0.9}{\small\input{./tables/result_drill_map.tex}    }
%     \scalebox{0.9}{\input{./tables/result_drill_map.tex}    }
%     \caption{\label{tab:drill_map} Human detection result: mAP on firefighter drill dataset.}
% \end{table}

% For qualitative analysis, the test data examples and corresponding detection results are shown in Fig. \ref{fig_map_abalation}. 
% The result proves that our UGF system can successfully work under challenging visual degradation.

% Based on the result, we believe there are many promising use cases of our system for emergency situations. 

% figure: field test data examples and corresponding detection results in the smoke-filled and dark environment.

\subsubsection{Runtime Efficiency}
Table~\ref{table_runtime} presents the model size and the inference speed of \sysname tested across various embedded devices, i.e., NVIDIA Jetson Xavier NX \cite{JetsonNX}, and NVIDIA Jetson AGX Xavier \cite{JetsonAGX}. The results show that \sysname has a similar model size and computation overhead when compared to the competing methods. Additionally, \sysname attains an inference speed of approximately 5 frames per second on Jetson Xavier NX and about 7 frames per second on Jetson AGX Xavier. Such latency is already able to support real-time inference and we believe the efficiency can be further improved by leveraging cutting-edge model compression methods (out of the scope of this work).

% The technical specifications of these embedding platforms can be found in Table \ref{tab:jetson_seriest}.

% \begin{table}[ht]
%     \centering
%     \scalebox{0.95}{ \footnotesize\input{./tables/Jetson_series} }
%     \caption{\label{tab:jetson_seriest} Technical specifications of different embedding platforms used for test.}
% \end{table}

% \usepackage{multirow}
\begin{table}[ht]
    \centering
    \caption{The model size and the inference latency of \sysname tested across various embedded devices.}
    \label{table_runtime}
    \scalebox{0.94}{
    \begin{tabular}{c|c|c|c|c} 
    \hline
    \multirow{2}{*}{Method} & \multirow{2}{*}{Paras} & \multirow{2}{*}{FLOPs} & \multicolumn{2}{c}{Frame Per Second on }                                                                                             \\ 
    \cline{4-5}
                            &                        &                        & \begin{tabular}[c]{@{}c@{}}Jetson\\Xavier NX\end{tabular} & \begin{tabular}[c]{@{}c@{}}Jetson\\AGX Xavier\end{tabular}  \\ 
    \hline
    VA \cite{chang2020spatial}                     & 7.24M                  & 61.4G                  & 4.94                                                      & 7.29                                                        \\
    AM \cite{li2022pedestrian}                     & 7.28M                  & 61.8G                  & 4.88                                                      & 7.44                                                        \\
                          Ours: \sysname  & 7.24M                  & 61.4G                  & 4.94                                                      & 7.21                                                        \\
    \hline
    \end{tabular}}
    \end{table}

%\begin{enumerate}
%    \item Experimental result 1 demonstrating performance A
%    \item Experimental result 2 demonstrating performance B
%    \item ...
%\end{enumerate}

%% file: tables/result_indoor_map.tex
\begin{threeparttable}
\begin{tabular}{l|lc|ccccc|ccccc} 
    \hline
    \multirow{2}{*}{Method} & \multicolumn{2}{c|}{Input} & \multirow{2}{*}{AP$_{50}$} & \multicolumn{1}{l}{\multirow{2}{*}{AP$_{65}$}} & \multicolumn{1}{l}{\multirow{2}{*}{AP$_{75}$}} & \multicolumn{1}{l}{\multirow{2}{*}{AP$_{95}$}} & \multicolumn{1}{l|}{\multirow{2}{*}{mAP$_{50:95}$ ↑}} & \multirow{2}{*}{mF1$_{50}$} & \multicolumn{1}{l}{\multirow{2}{*}{mF1$_{65}$}} & \multicolumn{1}{l}{\multirow{2}{*}{mF1$_{75}$}} & \multicolumn{1}{l}{\multirow{2}{*}{mF1$_{95}$}} & \multicolumn{1}{l}{\multirow{2}{*}{mmF1$_{50:95}$ ↑}}  \\ 
    \cline{2-3}
                            & \multicolumn{1}{c}{R\tnote{1}} & T\tnote{2}  &                                 & \multicolumn{1}{l}{}                                & \multicolumn{1}{l}{}                                & \multicolumn{1}{l}{}                                & \multicolumn{1}{l|}{}                                        &                                  & \multicolumn{1}{l}{}                                 & \multicolumn{1}{l}{}                                 & \multicolumn{1}{l}{}                                 & \multicolumn{1}{l}{}                                          \\ 
    \hline
    TOD \cite{Kristo2020Thermal}    &                       &   \checkmark  & 0.951                           & 0.828                                               & 0.524                                               & 0.001                                               & 0.527                                                        & 0.939                            & 0.843                                                & 0.642                                                & 0.008                                                & 0.583                                                         \\
    TOD-Radar \cite{Kristo2020Thermal} &         \checkmark               &    & 0.496                           & 0.286                                               & 0.102                                               & 0.000                                               & 0.194                                                        & 0.581                            & 0.408                                                & 0.229                                                & 0.002                                                & 0.277                                                         \\
    SOD \cite{liu2019salient}              &           \checkmark             &    \checkmark & 0.957                           & 0.879                                               & 0.661                                               & 0.001                                               & 0.588                                                        & 0.947                            & 0.884                                                & 0.734                                                & 0.013                                                & 0.635                                                         \\
    MilliEye \cite{shuai2021MilliEye} & \checkmark & \checkmark & 0.718 & 0.523 & 0.291 & 0.000 & 0.343 & 0.655 & 0.544 & 0.391 & 0.004 & 0.375 
    \\
    AM  \cite{li2022pedestrian}   &      \checkmark                  &    \checkmark & 0.952                           & 0.870                                               & 0.661                                               & 0.001                                               & 0.584                                                        & 0.935                            & 0.875                                                & 0.728                                                & 0.014                                                & 0.629                                                         \\
    VA   \cite{chang2020spatial}     &       \checkmark                &   \checkmark & 0.954                           & 0.886                                               & 0.681                                               & 0.001                                               & 0.594                                                        & 0.941                            & 0.886                                                & 0.743                                                & 0.010                                                & 0.635                                                         \\
    Ours: \sysname        &         \checkmark              &  \checkmark  & \textbf{0.962}\tnote{3}                  & \textbf{0.903}                                      & \textbf{0.764}                                      & \textbf{0.009}                                      & \textbf{0.644}                                               & \textbf{0.947}                   & \textbf{0.900}                                       & \textbf{0.795}                                       & \textbf{0.021}                                       & \textbf{0.672}                                                \\
    \hline
    \end{tabular}
    \begin{tablenotes}
        \footnotesize
        \item[1] denotes Radar depth image.
        \item[2] denotes Thermal image. 
        \item[3] \textbf{bold} denotes the best performance among all methods.
      \end{tablenotes}    
\end{threeparttable}        

%% file: sections/6_related_work.tex
\section{Related work}

\noindent \textbf{Object Detection with Thermal Cameras} 
The majority of thermal object detection works are dedicated to human detection and start with the adaptation from the established RGB-based object detection models. The authors in \cite{herrmann2018cnn} adopt the neural network models trained for RGB images and use transfer learning or cross-domain adaption to detect human subjects captured by thermal cameras. Wager et al. \cite{wagner2016multispectral} extend the amount of training data by utilizing the RGB red channel of the pure visual Caltech dataset to simulate additional thermal human detection data for more robust human detection. Guo et al. \cite{guo2019domain} achieved further improvement by combining real and synthetic training data. 
The authors in \cite{Ghose2019Pedestrain} tackle the situation that thermal images are less distinguishable due to the insufficient thermal contrast between people and their surroundings in the daytime. They propose to augment thermal images with their saliency maps to serve as an attention mechanism for pedestrian detection. In addition to human subject detection, the authors in \cite{Li2021YOLO-FIRI} utilise the YOLOv5 model architecture \cite{glenn_jocher_2022_7347926} for thermal detection on the FLIR dataset and the KAIST dataset. They modified the existing model architecture of the YOLOv5s by taking the CSP Bottleneck and applying an SK attention module. However, due to the limited contrast and textures of thermal images, the robustness of thermal-only solutions remains a question.

\noindent \textbf{Object Detection with mmWave Radars}
mmWave radars have been increasingly adopted in object detection tasks. These works can be categorised into radar-only solutions and radar-assisted multi-sensor fusion solutions. 
For radar-only solutions \cite{wang2021rodnet_radar, zheng2021scene}, the radio frequency (RF) data is input into neural networks to classify different types of objects on the road. \cite{meyer2021graph} leverages graphical representations of raw radar tensor data to gain a significant improvement in detection accuracy. \cite{danzer20192d, ulrich2022improved} use radar point cloud for vehicle detection. 
While the resolution of radar has been improved, the semantic information from the sparse and noisy radar data is noticeably challenging to extract and unsuitable for effective single-modality object detection. Radar-assisted multi-sensor fusion thus emerges to improve the detection robustness recently. 
For example, \cite{meyer2019deep, nabati2021centerfusion, hwang2022cramnet} present the camera-radar fusion architecture for accurate 3D object detection in safe autonomous driving. As mmWave radars are impervious to extreme weather conditions in which LiDAR falls short, \cite{li2022modality, qian2021robust} use radar-LiDAR fusion for vehicle detection. The radar-assisted multi-sensor fusion is also increasingly adopted in pedestrian or human detection. Authors in \cite{li2022pedestrian} propose a camera-radar fusion system for pedestrian liveness detection by exploiting the distinct reflection features of real pedestrians and distracting objects.
Taking advantage of the penetrability of radar signals and the ability to be unaffected by light, camera-radar fusion object detection system \cite{shuai2021MilliEye, deng2022global} is developed to detect humans in dark environments or through the fog. However, unlike \sysname, none of the above works systematically investigates the fusion of thermal cameras and mmWave radars for human detection. 

\noindent \textbf{Neural Sensor Fusion}
Based on the position of the fusion operation in a neural network, existing neural sensor fusion strategies can be categorized into three types: input fusion, feature fusion and decision fusion. Feature fusion is the most widely used in 
many areas: For disease diagnosis tasks, UncertaintyFuseNet \cite{abdar2023uncertaintyfusenet} and \cite{abdar2021barf} fuse cross-modal medical images, e.g., CT scans and X-ray images, by concatenating feature maps of two or more branches. For human activity recognition, RFCam \cite{chen2022rfcam} fuses Angle-of-Arrival, distance and activity features by proposed similarity scores. For object detection, the fusion between RGB cameras and other sensors, including radar sensors and LiDAR sensors, has been explored to improve detection performance: DeepFusion \cite{lee2020deep} and PointAugmenting \cite{wang2021pointaugmenting} augments points with their corresponding RGB features, \cite{nabati2021centerfusion} enriches RGB images with radar range measurement, CameraRadarFusionNet \cite{nobis2019deep}, RVNet \cite{john2019rvnet} and \cite{chadwick2019distant} fuses features of RGB branch and radar branch by channel concatenation. \cite{li2022pedestrian} and \cite{chang2020spatial} apply attention mechanism to fuse salient features. 
However, the above sensor fusion strategies usually cover only common sensors, e.g., RGB cameras, radar sensors and LiDAR sensors, while the novel thermal sensors are less explored. Besides, unlike RGB camera that outputs rich RGB information about the environment, both thermal images and radar images are weak in providing multi-channel measurements of objects (1-channel temperature information from thermal images and extremely sparse spatial location from radar images), it remains unknown whether and how thermal cameras and radar sensors can work together for optimum output.
% \cite{Devaguptapu2019Borrow}\cite{Roszyk2022Adopting} RGB+thermal fusion

% \begin{enumerate}
%     \item Related work in area A: RGB image-based human detection (representative work and its merits/limitation)
%     \item Related work in area B: thermal image-based human detection (representative work and its merits/limitation)
%     \item Related work in area C: mmWave human detection (representative work and its merits/limitation)
%     \item Related work in area D: Sensor fusion ..
% \end{enumerate}
% (Merge subsections if necessary)

% \textbf{SOTA methods for the target task}
% \begin{enumerate}
%     \item visual based method: Image-Adaptive YOLO for Object Detection in Adverse Weather Conditions
%     \item thermal based method: Thermal Object Detection in Difficult Weather Conditions Using YOLO
%     \item Adopting the YOLOv4 Architecture for Low-Latency Multispectral Pedestrian Detection in Autonomous Driving (RGB+Thermal)
%     \item MilliEye (RGB + mmWave radar, human detection, dark indoor environment) 
%     \item Seeing Through Fog Without Seeing Fog (multi-sensor fusion, object detection in adverse weather for autonomous driving)
%     %\item Learning to see through the haze: Multi-sensor learning-fusion System for Vulnerable Traffic Participant Detection in Fog (LiDAR+Radar, object detection in adverse weather for autonomous driving)
% \end{enumerate}

%% file: sections/7_conclusion.tex
\section{Conclusion}
By using thermal cameras and mmWave radars, this work presents a novel human detection approach \sysname as a robust alternative to RGB camera-based methods under visual degradation. \sysname utilizes a Bayesian feature extractor and an uncertainty-guided fusion method to systematically overcome the detection challenges caused by the low-resolution thermal images and the noisy radar point clouds. Our experimental results demonstrate that \sysname achieves superior performance compared to the state-of-the-art single-modal and neural sensor fusion methods. For future work, we plan to further improve the runtime efficiency of \sysname by using model compression techniques and generalize the uncertainty-guided fusion concept to other sensor combinations.